\pdfoutput=1

\documentclass[11pt]{article}

\usepackage[]{acl}

\usepackage{times}
\usepackage{latexsym}

\usepackage[T1]{fontenc}

\usepackage[utf8]{inputenc}

\usepackage{microtype}

\usepackage{graphicx}
\usepackage{booktabs}
\usepackage{amssymb}
\usepackage{amsmath}
\usepackage{longtable}
\usepackage{colortbl}
\usepackage{todonotes}
\usepackage{pifont}%

\newcommand{\xmark}{\ding{55}}%

\usepackage{xspace}
\usepackage{amstext}
\usepackage{multirow}
\usepackage{tabularx}   %
\usepackage{booktabs}   
\usepackage{arydshln}
\usepackage{subcaption}
\usepackage{colortbl}
\usepackage{xcolor}
\usepackage{svg}

\usepackage[most]{tcolorbox}

\usepackage[utf8]{inputenc} %
\usepackage[T1]{fontenc}    %
\usepackage{hyperref}       %
\usepackage{url}            %
\usepackage{booktabs}       %
\usepackage{amsfonts}       %
\usepackage{nicefrac}       %
\usepackage{microtype}      %
\usepackage{xcolor}         %

\newcommand\verythinrule{\specialrule{.02em}{0.3em}{0.2em}}

\usepackage{multirow}
\usepackage{arydshln}
\usepackage{graphicx}
\usepackage{booktabs}
\usepackage{amssymb}  %
\usepackage{subcaption}
\usepackage{amsmath} %

\usepackage{diagbox}
\usepackage{subcaption}

\usepackage{color, colortbl}

\usepackage{pifont}%

\usepackage{tabularx}
\usepackage{colortbl}
\usepackage{tikz} 
\usepackage{adjustbox}

\usepackage{xspace}
\newcommand\model{mmT5\xspace}
 \newcommand{\shared}{mT5$^S$\xspace}

\usepackage[utf8]{inputenc} %
\usepackage[T1]{fontenc}    %
\usepackage{hyperref}       %
\usepackage{url}            %
\usepackage{booktabs}       %
\usepackage{amsfonts}       %
\usepackage{nicefrac}       %
\usepackage{microtype}      %
\usepackage{xcolor}         %

\usepackage{multirow}
\usepackage{arydshln}
\usepackage{graphicx}
\usepackage{booktabs}
\usepackage{amssymb}  %
\usepackage{subcaption}
\usepackage{amsmath} %

\usepackage{diagbox}
\usepackage{subcaption}

\usepackage{color, colortbl}

\usepackage{pifont}%

\usepackage{tabularx}
\usepackage{colortbl}
\usepackage{tikz} 
\usepackage{adjustbox}

\usepackage{xspace}

\title{\model: Modular Multilingual Pre-Training\\Solves Source Language Hallucinations}

\author{%
  Jonas Pfeiffer \, \, \, Francesco Piccinno \, \, \, Massimo Nicosia \\ \textbf{Xinyi Wang \, \, \, Machel Reid \, \, \, Sebastian Ruder}\\
  Google DeepMind\\
}

\begin{document}

\maketitle

\begin{abstract}
Multilingual sequence-to-sequence models perform poorly with increased language coverage and fail to consistently generate text in the correct target language in few-shot settings.
To address these challenges, we propose \model, a modular multilingual sequence-to-sequence model. \model utilizes language-specific modules during pre-training, which disentangle language-specific information from language-agnostic information. We identify representation drift during fine-tuning as a key limitation of modular generative models and develop strategies that enable effective zero-shot transfer. Our model outperforms mT5 at the same parameter sizes by a large margin on representative natural language understanding and generation tasks in 40+ languages. Compared to mT5, \model raises the rate of generating text in the correct language under zero-shot settings from 7\% to 99\%, thereby greatly alleviating the source language hallucination problem.
\end{abstract}

\section{Introduction}

Multilingual pre-trained models \citep{conneau-etal-2020-unsupervised,xue-etal-2021-mt5} have demonstrated impressive performance on natural language understanding~(NLU) tasks across different languages \citep{hu2020xtreme,ruder-etal-2021-xtreme}. These models are typically trained on large amounts of unlabeled data in hundreds of languages. Recent large language models \citep{brown2020language,chowdhery2022palm} display surprising multilingual capabilities despite being pre-trained predominantly on English data. However, all of these models share a key limitation: representations of all languages compete for the model's limited capacity. As a result, models perform poorly with an increasing number of pre-training languages and on languages with less pre-training data. This is also known as the ``\textbf{curse of multilinguality}'' \citep{conneau-etal-2020-unsupervised}.

\begin{figure}[t]
    \centering 
        \includegraphics[width=.99\columnwidth]{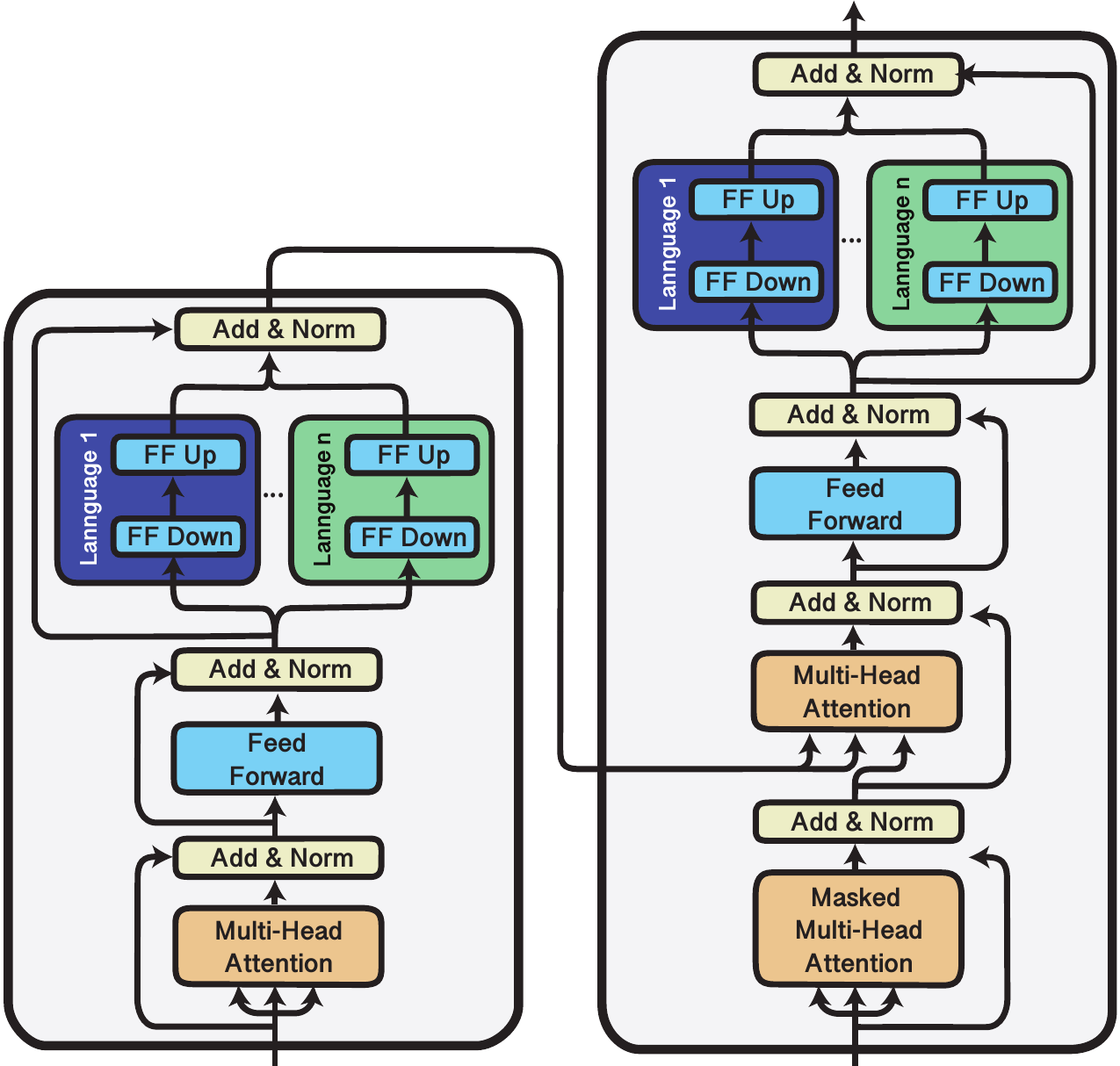}

    \caption{Architecture of \model. Language-specific bottleneck modules (dark blue and green components) are placed after the feed-forward component within each layer of the Transformer encoder-decoder model.
    }
\label{fig:mmt5}
\end{figure}

Natural language generation (NLG) tasks present another challenge for current multilingual models, which may overfit to the training languages and partially forget their generation ability in the target language \citep{vu-etal-2022-overcoming}, generating text with the \emph{correct} meaning in the \emph{wrong} language. We refer to this as the ``\textbf{source language hallucination problem}''.

To address these two limitations, we propose the modular multilingual T5 (\model, Figure~\ref{fig:mmt5}), the first modular multilingual generative model. During pre-training, \model~allocates a small amount of language-specific parameters to increase capacity for multilingual modeling. At fine-tuning time, we freeze the language-specific modules while tuning the shared parameters, allowing direct adaptation to a target language by swapping to the corresponding language-specific module. 

However, we observe an additional challenge for \model: the fine-tuned shared representations may drift away from the frozen modular representations in the decoder. The modular model is thus susceptible to generating text in the incorrect language, similar to its non-modular counterparts. To ameliorate this, we propose to freeze a subset of shared decoder parameters, which shows large improvements in zero-shot cross-lingual generation for modular generative models.

In general, we find that \model~is an effective model that overcomes the two limitations of multilingual sequence-to-sequence models: \textbf{1)} \model~alleviates the curse of multilinguality by adding additional model capacity to different languages during pre-training. It outperforms both standard baselines as well as mT5 \cite{xue-etal-2021-mt5} at the same parameter sizes on a representative set of multilingual NLU and NLG tasks; \textbf{2)}  \model~resolves the source language hallucination problem with impressive ability on zero-shot cross-lingual text generation. Our analysis~(\textsection\ref{subsec:halluciation}) shows that mT5 only generates text in the target language 7\% of the time for a zero-shot multilingual summarization task, while \model generates text in the correct language for 99\% of examples.

\section{Related work}

\paragraph{Modular language models} Much work has focused on \emph{post-hoc} modularity of pre-trained multilingual models, i.e., modular representations are added to existing dense models. The most commonly used modules are known as adapters \cite{Rebuffi2017adapters, Rebuffi2018, Houlsby2019adapters}. 
They enable specialization to new data settings  \cite{chen2019slice,Ruckle2020}, combination of new and existing knowledge \cite{Stickland2019BertPals, Wang2021KAdapter,pfeiffer-etal-2021-adapterfusion, Lauscher2020comonsense, Mahabadi2020Hyper, Poth2021Pretrain}, and adaptation to new cross-lingual \cite{pfeiffer-etal-2020-mad, pfeiffer2020unks,ustun-etal-2020-udapter, Vidoni2020OrthogonalLA, Ansell2021MADG, Ansell2021Composable, Wang2021Efficient} and NMT scenarios \cite{Bapna:2019emnlp, philip-etal-2020-monolingual, chronopoulou-etal-2020-reusing, Le:2021acl, Ustun2021DenoisingAda, Stickland2021DomainAdaNMT, garcia-etal-2021-towards, dua-etal-2022-tricks}. 

Our approach, in contrast, uses modularity \emph{a priori}, i.e., modularity is integrated into the module architecture as an inductive bias. Such modularity is similar to parameter sharing strategies commonly defined in multi-task learning \cite{ruder2017overview} as well as to mixture-of-experts approaches \cite[MoE;][]{Shazeer2017MoE}, which have been used to scale models to trillion parameters \cite{Fedus2021} and for domain-specific pre-training of LMs \cite{Gururangan2021Demix}. The most related work to ours is X-Mod \citep{pfeiffer-etal-2022-lifting}, which pre-trains an encoder-only BERT-style model in a modular fashion. Their model, however, cannot be used for natural language generation and underperforms our model on NLU tasks (see Section  \ref{sec:experiments}).

\paragraph{Limitations of multilingual language models} State-of-the-art multilingual LMs are pre-trained on large amounts of multilingual data in around 100 languages. Prior work has demonstrated, however, that models' performance deteriorates with increasing language coverage given the same fixed capacity, known as the \textit{curse of multilinguality} \citep{Conneau2020xlm-r}. Prior studies also found that models perform poorly on languages that are under-represented in pre-training \cite{wu-dredze-2020-languages,hu2020xtreme,Lauscher:2020zerohero,artetxe-etal-2020-cross,pfeiffer-etal-2020-mad,pfeiffer2020unks,ChauLS20Parsing,ponti-etal-2020-xcopa}. For natural language generation, multilingual models have been observed to overfit to the source language and fail to generate text consistently in the correct target language \cite{vu-etal-2022-overcoming}.

\section{mmT5}
Standard multilingual models update the same model parameters for hundreds of languages during pre-training, resulting in the curse of multilinguality where different languages compete for the limited model capacity~\citep{conneau-etal-2020-unsupervised}. We propose \model, the first modular sequence-to-sequence multilingual model that allocates language specific modules during pre-training. In this section, we discuss the architecture of \model, its training and fine-tuning methods, and our strategies to resolve the source language hallucination problem with \model.

\subsection{Modeling}
First, we describe the overall architecture of \model. We augment a standard Transformer encoder-decoder model with language-specific  modules at every transformer layer (see Figure~\ref{fig:mmt5}). The selection of modules \cite[i.e., fixed routing;][]{pfeiffer-etal-2023-modulardeeplearning} is performed via the language ID provided with each  example\footnote{Our pre-training data contains such metadata; alternatively, automatic language ID methods can be used (see \S\ref{subsec:halluciation}).}; all tokens of an example are passed through the same language-specific module. 

We use bottleneck adapters as the language-specific module because they perform better at smaller model sizes compared to other modular methods such as continuous prompts \citep{karimi2021compacter,He2022towards}. We place a module after the feed-forward component in each layer. In contrast to \citet{pfeiffer-etal-2022-lifting} that only experimented with encoder-only models, we focus on a more general sequence-to-sequence model following the T5 architecture \cite{raffel2020exploring}.

We add $N \times L$ modular components to the T5 architecture where $L$ is the number of layers of the model and $N$ corresponds to the number of languages which the model is pre-trained on. The transformer weights are shared across languages while the modular component provides the model with language-specific capacity. During a forward pass, each input is first passed through the shared transformer weights and then routed through the corresponding language-specific module based on the language of the input. 
We follow this procedure for all transformer layers until the representations are passed to the shared prediction head.

\subsection{Modular Pre-training, Fine-tuning, and Inference}
We pre-train both language-specific modules and shared parameters jointly. During fine-tuning, we freeze all language-specific modules and only update the shared parameters. This paradigm allows us to more effectively adapt the fine-tuned model to any of the languages included in the pre-training data by simply switching to the corresponding language-specific module. At inference, the module corresponding to the target language is used together with the fine-tuned shared parameters.

\subsection{Overcoming Modular Representation Drift} \label{sec:modular-representation-drift}

When fine-tuning the modular model for transfer settings in \textsection\ref{sec:results}, we observe a scenario of \emph{modular representation drift}: we find that the shared parameters that are updated during task-specific training drift away from the modular parameters and become thus less compatible with modules that are used for inference. In practice, this leads to a loss of compositional generalization where the modular model generates text in the incorrect language, similar to its non-modular counterparts \cite{vu-etal-2022-overcoming}; see \S\ref{subsec:halluciation}.

In order to ameliorate this drift, we propose to freeze parts of the model, with a focus on the decoder. We find that freezing the decoder feed-forward parameters provides the biggest benefit (see \textsection\ref{sec:freezing-analysis} for the detailed ablation) and almost completely eliminates the source language hallucination problem in modular models.\footnote{We observe little benefit to freezing decoder parameters in non-modular models, however.}

\section{Experiments} \label{sec:experiments}

\paragraph{Pre-training Details}
We pre-train mmT5 on data from  100 languages in mC4 \cite{xue-etal-2021-mt5} following the general pre-training setup of mT5 \cite{xue-etal-2021-mt5}, if not specified otherwise. We pre-train mmT5 at two model sizes: small (300M parameters), and base (580M parameters).
We train model variants with an input sequence length of 1024 and a target sequence length of 256 for 1M update steps with a batch size of 1024.
The bottleneck size of each module is half of the hidden dimension of the transformer model. For instance, as the base variant has a hidden dimension of 768, we set the bottleneck size to 384.\footnote{We analyze the impact of bottleneck sizes in \S\ref{sec:analysis-bottleneck-size}.} We additionally pre-train a non-modular variant of our modular model, \shared, where all parameters are shared across all languages. The \shared variant uses exactly the same hyper-parameters and pre-training setup as \model. To ensure that the models are directly comparable and have exactly the same number of parameters, we add shared bottleneck layers to \shared in the same configuration as in \model.

\paragraph{Experimental setting}

We conduct experiments across datasets in \textit{zero-shot cross-lingual transfer} and \textit{multilingual training} scenarios. For \textit{zero-shot cross-lingual transfer}, we train the model on a subset of languages (e.g., only English) and evaluate the model on held-out data of the same task in other languages. In \textit{multilingual training}, we fine-tune the model on multiple languages of the same task, and evaluate the model on the same set of languages. As the language-specific modular components are replaced at inference time, we do not update the parameters of the modular components (i.e., we freeze the modules). We do the same for our shared model variants, in order for the number of trainable parameters to be equal for comparable scenarios.\footnote{Here, we follow the procedure of \citet{pfeiffer-etal-2022-lifting}.} For each dataset, we select the best model checkpoint based on performance on the validation set.

\paragraph{Evaluation Tasks}
For \textit{zero-shot cross-lingual transfer}, we evaluate on the XQuAD \cite{artetxe-etal-2020-cross} and TyDi QA GoldP \cite{clark-etal-2020-tydi} question answering datasets; on the XNLI \cite{conneau-etal-2018-xnli} natural language inference dataset; on XL-Sum \cite{hasan-etal-2021-xl} for summarization;\footnote{We do not evaluate on Oromo, Kirundi, Pidgin, and Tigrinya as they were not seen during pre-training.} 
and MASSIVE \cite{fitzgerald2022massive} for semantic parsing.\footnote{We do not evaluate on Hebrew and Tagalog as they were not seen during pre-training.} We mainly fine-tune the model on English training data and evaluate on the target languages \cite{hu2020xtreme}. For XL-Sum, we additionally evaluate in a multi-source zero-shot transfer setting where we train jointly on data in Arabic, English, Japanese and Chinese (XL-Sum$^{ar,en,ja,zh}$).

For \textit{multilingual training}, we evaluate on semantic parsing (MASSIVE) and summarization (XL-Sum) datasets. For each dataset, we fine-tune and evaluate the model on all languages jointly.

\begin{table}[]
\centering
\resizebox{\columnwidth}{!}{%
\begin{tabular}{lccc}
\toprule
\multirow{2}{*}{Model} & \multirow{2}{*}{Variant} &  \multirow{2}{*}{Shared Params.} & Mod. Params. \\
& & & per Lang. \\ \midrule
mBERT & Base & 178M & --\\ \verythinrule
X-Mod & Base & 270M & 7M\\ \verythinrule
\multirow{2}{*}{XLM-R} & Base & 270M & --\\ 
 & Large & 550M & --\\ \verythinrule
\multirow{2}{*}{mT5} & Small & 300M & --\\
 & Base & 580M& -- \\
\multirow{2}{*}{mT5$^S$} & Small & 300M + 4M & -- \\
 & Base & 580M  + 14M & -- \\
\verythinrule
\multirow{2}{*}{mmT5} & Small & 300M & 4M  \\
 & Base & 580M  & 14M \\
\bottomrule
\end{tabular}%
}
\caption{Number of shared and modular parameters of baselines and our models.}
\label{tab:model_sizes}
\end{table}

\begin{figure}[t]
    \centering 
        \includegraphics[width=.99\columnwidth]{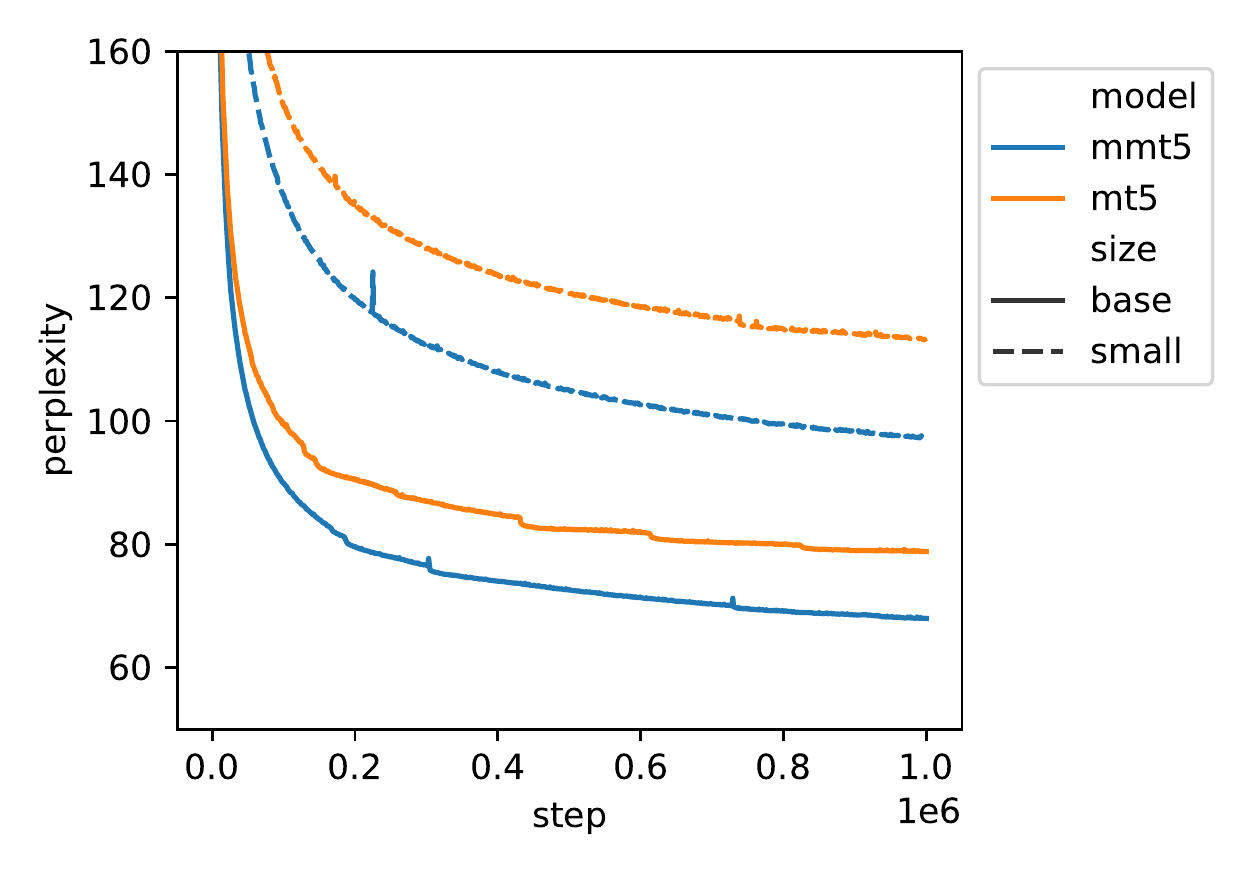}

    \caption{Perplexity (lower is better) of different model sizes during pre-training for \model and \shared, averaged across languages.
    }
\label{fig:perplexity}
\end{figure}

\paragraph{Baselines} Our main comparison method is \textbf{\shared}, a shared model that is pre-trained with the same hyper-parameters, setup, and number of parameters as our modular model. We also compare to the published results of the \textbf{mT5} encoder-decoder model \cite{xue-etal-2021-mt5}. In addition, we compare to several encoder-only models including \textbf{mBERT} \cite{Devlin2019bert}, \textbf{X-Mod} \cite{pfeiffer-etal-2022-lifting}, and \textbf{XLM-R} \cite{Conneau2020xlm-r}. Encoder-only models are generally smaller as they lack a decoder but cannot easily be used for generation tasks. We provide an overview of the model sizes of the baselines and our method in Table \ref{tab:model_sizes}.

\paragraph{Decoder Freezing Configurations}

To overcome the modular representation drift described in \textsection\ref{sec:modular-representation-drift}, we experiment with different configurations of freezing parts of the model when fine-tuning the model on a downstream task. We experiment with freezing the LayerNorm (LN), self-attention (Att), cross-attention (CrossAtt) and feed-forward component (FFN) in the encoder (Enc) and decoder (Dec) parts of the transformer model. We ablate freezing configurations in \textsection\ref{sec:freezing-analysis} and report test results of the freezing configuration that performs best on the dev set for each dataset for \model. For dense models, we observe no impact with freezing and report results using full fine-tuning.

\section{Results}
\label{sec:results}

\begin{table*}[]
\begin{center}
\def\arraystretch{0.87}
\resizebox{0.99\textwidth}{!}{%
\addtolength{\tabcolsep}{-0.5pt}
\begin{tabular}{lllcccccc}
\toprule
&    \multicolumn{2}{c}{\textit{Zero-Shot}}     &  XQuAD  & TyDiQA(GoldP)    & XNLI& XL-Sum$^{en}$ & XL-Sum$^{ar,en,ja,zh}$ & MASSIVE \\
                  &    &             & F1 / EM   & F1 / EM    & Acc& RG$_1$ / RG$_2$ / RG$_L$ & RG$_1$ / RG$_2$ / RG$_L$ & EM \\
\midrule
\parbox[t]{2mm}{\multirow{4}{*}{\rotatebox[origin=c]{90}{Encoder}}} & \multirow{3}{*}{base} & mBERT & 64.5 / 49.4 & 59.7 / 43.9 & 65.4 & --- & --- & ---  \\
&  & X-Mod & 72.8$^*$ / --- & --- / --- & 73.5$^*$  & ---  & --- & ---  \\
&  & XLM-R & 70.6 / 55.5 & --- / --- & 76.2 & ---   & --- & ---     \\
\cmidrule{2-9}
& large & XLM-R & 76.6 / 60.8 & 65.1 / 45.0 & 79.2 &   ---  & --- & ---   \\
\midrule
\parbox[t]{2mm}{\multirow{6}{*}{\rotatebox[origin=c]{90}{Encoder-decoder}}} & \multirow{3}{*}{small} & mT5         & 58.1 / 42.5 &  35.2 / 23.2 & 67.5 & ---    & ---  & ---    \\
&                        & \shared  & 61.9 / 46.2 & 44.5 / 31.1  &  63.2  & 15.5 / \,	2.2 /	14.2 & 17.0 / \,	4.7 /	15.1  & 21.7  \\
&                        & \model & \textbf{66.5 / 50.4} & \textbf{50.8 / 36.3} &  \textbf{68.5}   & \textbf{16.7} / \,	\textbf{4.6} /	\textbf{14.4} & \textbf{29.4} /	\textbf{12.6} /	\textbf{23.3}  &  \textbf{27.7}  \\
\cmidrule{2-9}
& \multirow{3}{*}{base}  & mT5         & 67.0 / 49.0 & 59.1 / 42.4 & 75.4 &  ---     & ---  & 34.7        \\
&                        & \shared  & 68.7 / 51.5 & 64.0 / 47.8 & 75.1 &  16.2 / \,	2.8 / \, 4.5 & 18.6 / \,	6.0 / 16.7 & 39.9    \\
&                        & \model & \textbf{76.3	/ 60.3} & \textbf{69.0 / 53.2} & \textbf{77.8} & \textbf{19.6} / \, \textbf{6.1} /	\textbf{16.4}  & \textbf{34.5} / \textbf{16.1} / \textbf{26.8} & \textbf{46.0}   \\
\bottomrule
\end{tabular}   
}
\end{center}
\caption{Zero-shot cross-lingual transfer test results averaged over all languages. mBERT and XLM-R scores are from \cite{hu2020xtreme}; XLM-R Base XNLI results are from \cite{Conneau2020xlm-r}; mT5 results are from \cite{xue-etal-2021-mt5}; X-Mod results are from \cite{pfeiffer-etal-2022-lifting} (${}^*$average is only on a subset of languages).
}
\label{tab:main-test-zero-shot-results-average-languages}
\end{table*}

\begin{table}[]
\begin{center}
\def\arraystretch{0.87}
\resizebox{0.99\columnwidth}{!}{%
\addtolength{\tabcolsep}{-0.5pt}
\begin{tabular}{lllcc}
\toprule
      \multicolumn{2}{c}{\textit{Multilingual}}     &    &               XL-Sum & MASSIVE \\
                  &    &   &   RG$_1$ / RG$_2$ / RG$_L$ & EM \\
\midrule
\parbox[t]{2mm}{\multirow{4}{*}{\rotatebox[origin=c]{90}{Enc-dec}}} & \multirow{2}{*}{small}
       &  \shared           & 36.4 /  17.9 / 28.5 & 60.7     \\
&                        & \model     &    \textbf{36.7} / \textbf{18.1}  / \textbf{28.7} & \textbf{65.6}   \\
\cmidrule{2-5}
& \multirow{2}{*}{base}  
                        & \shared   &   39.1 / 20.3 / 30.5 & 64.6   \\
&                        & \model    &  \textbf{41.6} / \textbf{22.8} / \textbf{33.0} & \textbf{66.7}  \\
\bottomrule
\end{tabular}   
}
\end{center}
\caption{Multilingual training test results averaged over all languages.
}
\label{tab:main-test-multi-source-results-average-languages}
\end{table}

\subsection{Pre-training}

We first compare the language modeling perplexities of different model sizes for \model~and \shared~during pre-training in Figure~\ref{fig:perplexity}. We find that \model~significantly outperforms its fully shared counterpart during the early stages of pre-training and maintains the gap throughout the pre-training process. From an efficiency perspective, \model only requires 282k and 220k update steps respectively for the small and base versions to achieve the same final perplexity as the \shared models at 1M update steps. This corresponds to a $\approx 4 \times$ efficiency boost when training a modular multilingual model compared to a fully dense one.   

\subsection{Fine-tuning} 

We present our main results on the test sets for zero-shot cross-lingual transfer and multilingual training scenarios  in Tables~\ref{tab:main-test-zero-shot-results-average-languages} and~\ref{tab:main-test-multi-source-results-average-languages}, respectively.
\model outperforms both the original mT5 as well as \shared across all model sizes. It achieves performance similar to XLM-R at the same parameter size---despite its encoder-decoder configuration---and significantly outperforms X-Mod, the only other modular model.

\paragraph{Zero-shot} 
For zero shot cross-lingual transfer scenarios, we see large gains for generative tasks in particular. For question answering (XQuAD and TyDiQA), we observe an average relative F1 improvement of 5.5 and 6.3 for the small and base models respectively. For summarization, we see larger zero-shot gains when jointly training on more than one language. We suspect that this is due to the increase in training data and due to positive transfer during multi-source training, which modular methods are better able to harness. This is in line with previous findings that multi-source training improves cross-lingual transfer in adapter-based setups \cite{ansell-etal-2021-mad-g}. We also see a gain of 6.1 EM points on MASSIVE. The smallest gains are achieved for the classification task XNLI. Here, \model improves over the baselines only by 1--2.4 accuracy points. We hypothesize that due to the constrained formulation of the task, which only requires predicting a single token, the full multilingual generative capabilities of \model are under-utilized. Overall, we see a clear trend that our modular models significantly outperform their respective dense counterparts especially for generation tasks. 

\paragraph{Multilingual training} For multilingual training in Table \ref{tab:main-test-multi-source-results-average-languages}, we also find that the modular models outperform their dense counterparts across all tasks we experiment with. Here we find the largest gains for semantic parsing (MASSIVE). For summarization (XL-SUM), we see smaller, but still consistent gains. These results indicate that modular representations are not only useful in transfer settings but that \model~can also leverage labeled data in the target language to deliver superior performance compared to the standard non-modular models.

\begin{table*}[]
\begin{center}
\def\arraystretch{0.87}
\resizebox{0.99\textwidth}{!}{%
\addtolength{\tabcolsep}{-0.5pt}
\begin{tabular}{cccccclcccclc}
\toprule
 &  &  &  &  &  &  & \multicolumn{4}{c}{Zero-Shot} &  & \multicolumn{1}{c}{Multilingual} \\
 &  &  &  &  &  &  & \multicolumn{2}{c}{XQuAD} & XNLI & MASSIVE &   & XL-Sum \\
 &  &  &  &  &  &  & dev (en) & test & dev & dev &   & dev \\
Emb & Enc$_\text{LN}$ & Dec$_\text{LN}$ & Dec$_\text{Att}$ & Dec$_\text{CrossAtt}$ & Dec$_\text{FFN}$ &  & f1 / em & f1 / em & acc & EM &   & Rg$_1$ / Rg$_2$ / Rg$_L$ \\ \midrule
 &  &  &  &  &  &  & 90.7 /  83.6 & 66.9 /  49.3 & 75.5 & 32.1 &    & 41.2 / 22.4 / 32.4 \\
\xmark &  & \xmark &  & \xmark & \xmark &  & 91.9 /  85.1 & 75.8 /  59.5 & 75.6 & 43.2 &    & 41.2 / 22.4 / 32.6 \\
\xmark &  & \xmark &  &  & \xmark &  & 91.8 /  85.1 & 75.8 /  59.8 & 77.3 & 41.0 &   & \textbf{41.9} / \textbf{23.1} / \textbf{33.2} \\
\xmark & \xmark & \xmark &  & \xmark & \xmark &  & \textbf{92.1} /  \textbf{85.5} & \textbf{76.3} /  \textbf{60.3} & 76.1 & \textbf{45.4} &    & 40.8 / 22.1 / 32.3 \\
\xmark & \xmark & \xmark &  &  & \xmark &  & 91.8 /  85.1 & 75.0 /  59.2 & \textbf{77.7} & 39.9 &    & 41.8 / 23.0 / 33.1 \\
\bottomrule
\end{tabular}
}
\end{center}
\caption{Results of different freezing configurations for \model base on different tasks. Dev results for most. We always fine-tune Enc$_{Att}$, and Enc$_{FFN}$ and always freeze Enc$_{Mod}$ and Dec$_{Mod}$. \xmark \,  indicates that this component is frozen during task-level fine-tuning.
}
\label{tab:config-setup-averaged-dev-results}
\end{table*}

\section{Analysis and Ablations} \label{sec:analysis}

\subsection{Impact of Freezing Configuration} \label{sec:freezing-analysis}

We investigate the impact of the freezing configuration on the performance of the model. In Table~\ref{tab:config-setup-averaged-dev-results}, we compare the best-performing freezing configurations with a non-frozen baseline for \model base (we show the results of all freezing configurations in Appendix \ref{sec:freezing_combinations}). We observe significant improvements when freezing the feed-forward layer of the decoder during fine-tuning, particularly in zero-shot scenarios. For multilingual training, freezing of the decoder has less effect on the performance. We also find that freezing parts of the decoder has no effect on the dense \shared model across all tasks (see Appendix \ref{sec:freezing_combinations}).

\begin{figure}[]
    \centering
        \includegraphics[width=.99\linewidth]{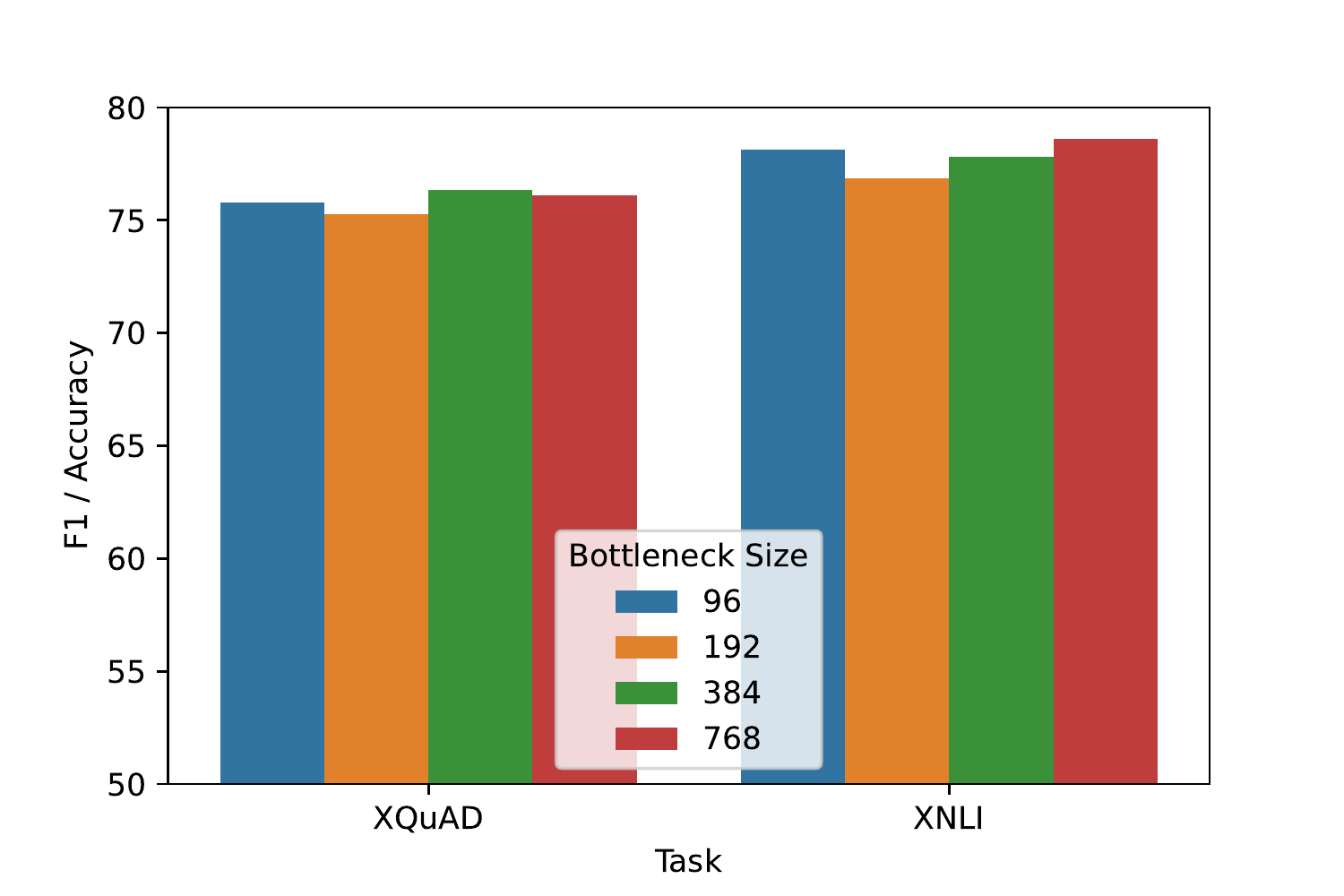}

    \caption{Comparison of bottleneck sizes of base \model for XQuAD (F1) and XNLI (Accuracy). 
    }
\label{fig:bottleneck_sizes}
\end{figure}

\subsection{Impact of Bottleneck Size} \label{sec:analysis-bottleneck-size}

We experiment with different bottleneck sizes of the modular components to understand the impact of providing each language with more capacity. We report results for XQuAD, and XNLI in Figure~\ref{fig:bottleneck_sizes} using \model base and bottleneck sizes of 96, 192, 384, and 768. We find that for all three tasks the bottleneck size has little effect on the downstream task performance, achieving only 0.5--2 absolute points difference between the larger and the smaller bottleneck sizes. This suggests that it is sufficient to provide the model with only a small amount of language-specific  parameters in order to learn idiosyncratic information and mitigate catastrophic interference, and highlights the parameter-efficiency of modular models.

\subsection{Impact of Model Size}

\begin{figure}[]
    \centering
        \includegraphics[width=.99\linewidth]{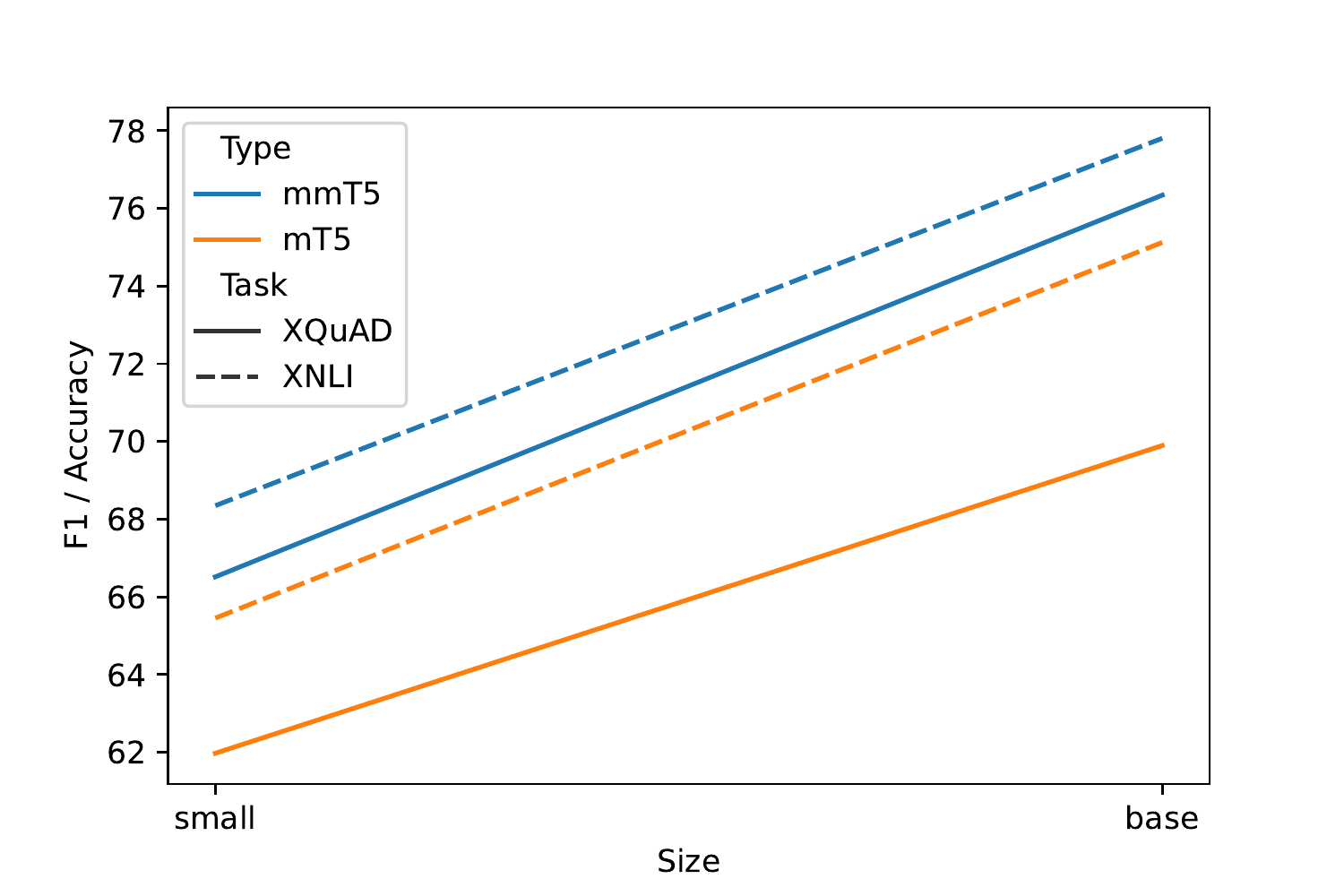}
    \caption{Comparison of model sizes  for XQuAD (F1) and XNLI (Accuracy).
    }
\label{fig:size_comparison}
\end{figure}

\begin{figure*}
\begin{subfigure}[b]{0.5\textwidth}
\begin{tcolorbox}[nobeforeafter, colback=white]
Südkalifornien besteht aus [\ldots]
einer internationalen Metropolregion und Großstadtgebieten. Die Region ist die Heimat von \colorbox{green}{zwei} erweiterten Metropolregionen mit jeweils mehr als fünf Millionen Einwohnern. [$\ldots$]
\\
\textbf{Question}: 
Wie viele erweiterte Metropolregionen gibt es ? 
\tcblower
\textbf{\model}: \colorbox{green}{zwei} \\
\textbf{\shared}: \colorbox{red}{two}
\end{tcolorbox}
\label{fig:xquad_example_1}
\end{subfigure}
\hfill
\begin{subfigure}[b]{0.5\textwidth}
\begin{tcolorbox}[nobeforeafter, colback=white]
[\ldots]
Analysen [\ldots]
waren irreführend, da es \colorbox{green}{mehrere Jahre} dauert, bis die Auswirkungen zu Veränderungen des Wirtschaftswachstums führen. [\ldots] \\
\textbf{Question}: 
Wie lange dauert es, bis sich die Auswirkungen als Veränderungen des wirtschaftlichen Wachstums manifestieren? 
\tcblower
\textbf{\model}: \colorbox{green}{mehrere Jahre} \\
\textbf{\shared}: \colorbox{red}{more}\colorbox{green}{ere Jahre}
\end{tcolorbox}
\label{fig:xquad_example_3}
\end{subfigure}
\vspace{-2.5em}
\caption{XQuAD examples where \shared generates tokens with the correct meaning but in the wrong language. For the same examples, \model is able to generate tokens in the correct language when freezing parts of the decoder.}
\label{fig:xquad_examples}
\end{figure*}

In Figure~\ref{fig:size_comparison}, we plot the performance difference of \model and \shared for the small and base variants. We find that the modular model outperforms the dense variant across model sizes with a similar gap, indicating that the positive effect of modularity may not diminish at scale.

\subsection{Source Language Hallucination\label{subsec:halluciation}}

\begin{figure}[]
    \centering
        \includegraphics[width=.99\linewidth]{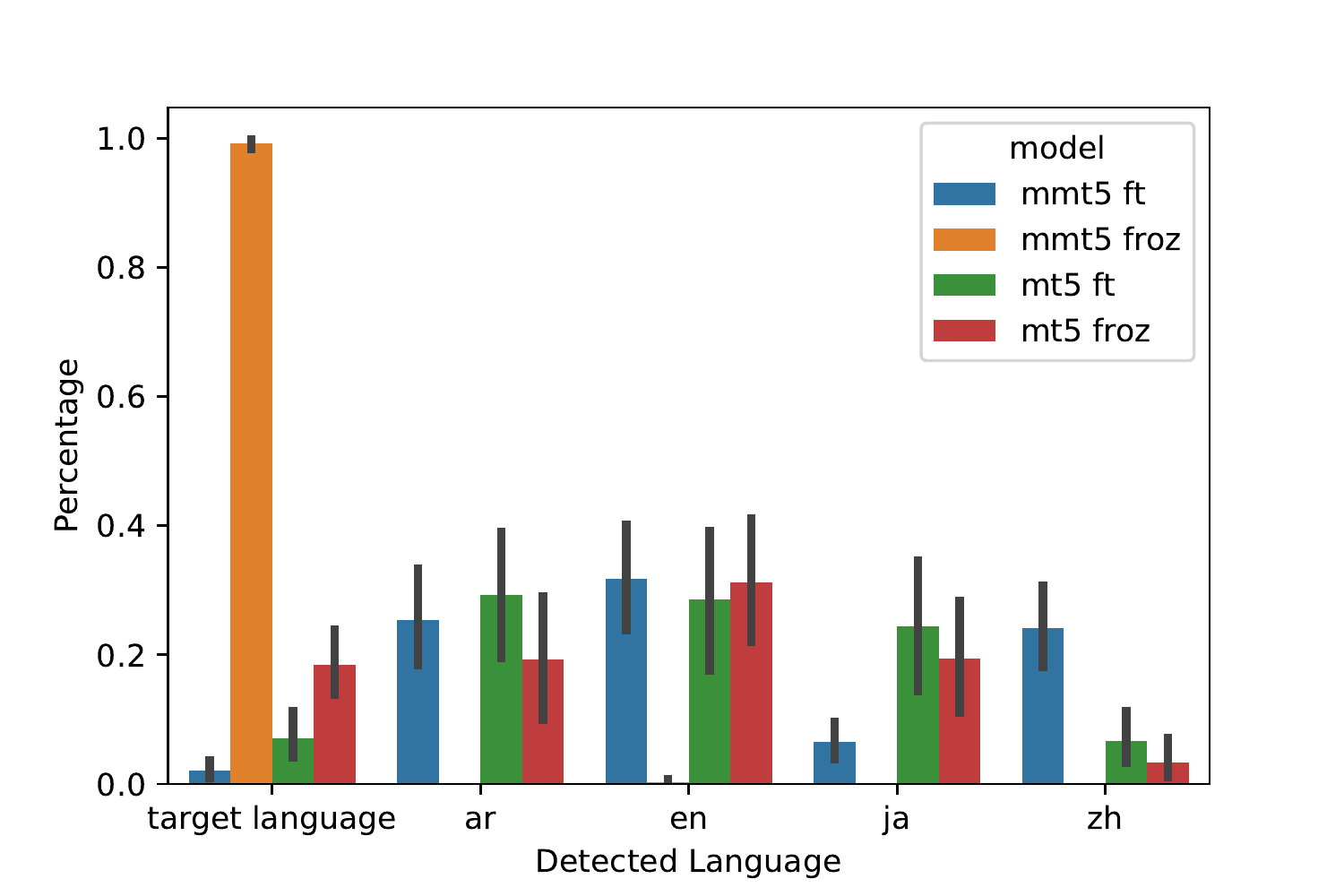}
    \caption{Detected languages of generated text of the development set of XL-Sum$^{ar,en,ja,zh}$. All models have base size. *ft indicates that the decoder was finetuned, *froz indicates that the decoder was partially frozen. High numbers are desireable  for the first set of plots (``target language''), low numbers are desireable for the remaining four sets of plots (``ar'', ``en'', ``ja'', ``zh''). We only include zero-shot cross-lingual results, therefore exclude the four source languages; all models achieve 100\% accuracy for those. For more granular results see Appendix Table~\ref{tab:lang_detection_1}.
    } 
\label{fig:lang_id} 
\end{figure}

We perform an analysis of the generated text on the XL-Sum dev sets for \shared and \model models trained in a zero-shot setting on XL-Sum$^{ar,en,ja,zh}$ using full fine-tuning and a decoder freezing configuration. We automatically detect the language of the generated text using the Language Detection from the Google Cloud Translation API\footnote{\url{https://cloud.google.com/translate/docs/basic/detecting-language}} \cite{caswell-etal-2020-language}. We show the results in Figure~\ref{fig:lang_id}. We find that most models tend to generate text in one of the source languages (in this setting: Arabic, English, Japanese, and Chinese). This holds true also for \model when we fine-tune the decoder. However, when freezing the decoder we observe a dramatic improvement in the target language generation rate from 1\% to 99\% of examples for \model, essentially solving the issue of source language hallucination in cross-lingual transfer scenarios. This improvement in language consistency also helps explain the significant improvement of the modular model over its dense counterparts on natural language generation tasks.

In addition, we manually analyze outputs of \shared and \model on XQuAD and find similar issues of source language hallucinations. We show examples in Figure \ref{fig:xquad_examples}. Although the task is extractive QA, i.e., the answer is a substring of the input, \shared tends to translate subwords into English (the source language). This does not happen to \model when freezing parts of the decoder, partially explaining the large improvements of \model over \shared on TyDi QA in Table~\ref{tab:main-test-zero-shot-results-average-languages}.

\begin{figure*}[]
    \centering
        \includegraphics[width=1\textwidth]{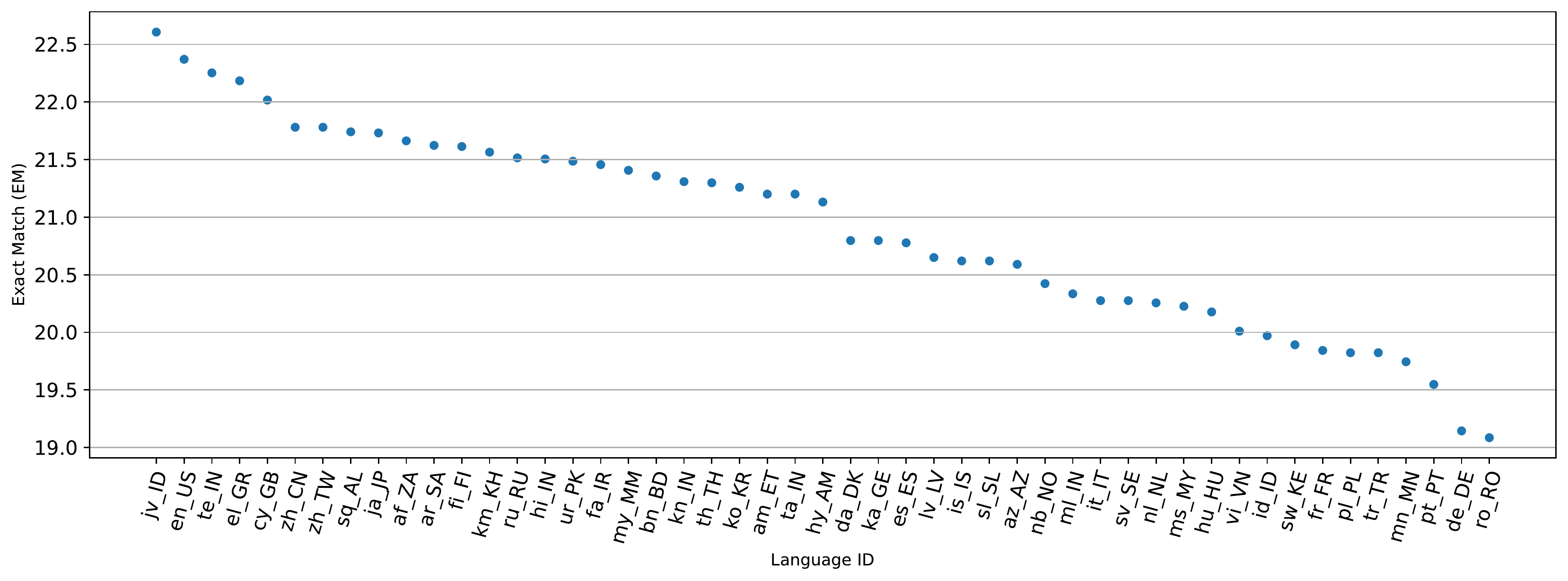}
    \caption{Average of the top-5 zero-shot EM accuracies on the Tagalog MASSIVE development set by varying the input language ID. Tagalog was not seen during mmT5 pre-training.} 
\label{fig:tagalog} 
\end{figure*}

\subsection{Module Re-Use for Unseen Languages}
\label{sec:unseen_languages}

In the previous sections we have evaluated the cross-lingual performance of \model on languages seen during pre-training. However, with more than 7000 languages spoken in the world \cite{joshi2021}, \model covers less than 1\% of them. While extending the model to unseen languages is out of scope for this work\footnote{Previous work has demonstrated that it is possible to extend multilingual models to unseen languages \cite{pfeiffer-etal-2020-mad, pfeiffer2020unks, pfeiffer-etal-2022-lifting}.}, we evaluate the potential reusability of existing language modules for truly unseen languages with a case study on Tagalog. 
We utilize the  base \model model fine-tuned on the English MASSIVE  training dataset (see Table~\ref{tab:main-test-zero-shot-results-average-languages}). As a Tagalog language module does not exist within \model,  we test all existing \textit{other} language modules when evaluating on the Tagalog test set. In Figure~\ref{fig:tagalog}, we report the Exact Match (EM) zero-shot accuracies for all languages.
The module performing best corresponds to Javanese, which is the most closely related language to Tagalog as both belong to the Malayo-Polynesian subgroup of the Austronesian language family.
This finding demonstrates the effectiveness of modular models; modular components specifically incorporate interpretable concepts, which can be re-used for unseen scenarios. 
Additionally, they can be further fine-tuned or adapted to the target domain if training data is available.

\section{Conclusion}
We have proposed \model, a modular multilingual encoder-decoder model. During multilingual pre-training the majority of parameters of \model are shared between tasks, but each language is provided with a small amount of parameters only accessible to the respective language. We demonstrated that integrating modularity as an architectural inductive bias significantly improves training efficiency, where the same perplexity as an equivalent fully dense model is achieved at a quarter of the update steps. \model considerably outperforms comparable models on a large number of tasks including Question Answering, Semantic Parsing, Summarization and Classification in both zero-shot as well as multilingual scenarios. Finally, we show that by freezing parts of the decoder when fine-tuning \model on a target task in a source language, the model consistently generates text in the target language. Consequently, modularity arguably solves source language hallucinations in cross-lingual transfer scenarios.  

\section{Future Work}
In this paper, we explored the use of modularity for multilingual language models. We showed that modularity significantly improves cross-lingual performance on a number of generative tasks by mitigating hallucinations in the source language. 
However, there are still many avenues for future work.

First, we did not consider placing the modules in different parts of the model. We only experimented with placing bottleneck layers after the feed-forward component of each transformer layer. Previous work has demonstrated that depending on the modality, different placements perform better \cite{Pfeiffer2020adapterfusion, eichenberg-etal-2022-magma}.

Second, we only experimented with extending the vanilla transformer architecture with modular components. Future work might consider modularizing different parts of the transformer, such as the attention-components or entire feed-forward layers like in \citet{Kudugunta2021TaskMOEs}.  

Third, we performed fixed routing under the assumption that the language ID is easy to obtain. We chose this path, as learning-to-route has many difficulties such as training instabilities \cite{pfeiffer-etal-2023-modulardeeplearning}. However, this architecture design limits the sharing of information (e.g. domains) across languages. Consequently, a combination of fixed routing and learned routing  would allow the model to learn how to share information across subsets of languages.

Fourth, we did not try using \model for machine translation. Using a modular design for this type of task setup is quite natural, as modules from the encoder and decoder can be easily replaced with the source and target language components, respectively. The effectiveness of modular sequence-to-sequence models for NMT has been investigated previously \cite{Bapna:2019emnlp, philip-etal-2020-monolingual, chronopoulou-etal-2020-reusing, Le:2021acl, Ustun2021DenoisingAda, Stickland2021DomainAdaNMT, garcia-etal-2021-towards, dua-etal-2022-tricks}. 

Finally, we did not consider extending the model to languages beyond those we pre-trained on. While our preliminary results (see \S~\ref{sec:unseen_languages}) suggest that there are benefits of reusing related language modules to learn unseen languages, this requires further experimentation. However, previous works have demonstrated that modular \cite{pfeiffer-etal-2022-lifting} as well as  dense models can be adapted to new languages and scripts \cite{pfeiffer-etal-2020-mad,pfeiffer2020unks}. Alternatively, future work might consider using post-hoc adaptation techniques, such as LoRA \cite{Hu2022LoRA}, to adapt modules to new languages.

\section*{Acknowledgements}

We thank Andrea Gesmundo, Marc'Aurelio Ranzato and Srini Narayanan for helpful feedback on a draft of this paper.

\footnotesize{
\bibliography{anthology,main}
\bibliographystyle{acl_natbib}
}

\appendix

\section{Appendix}

\subsection{Freezing combinations}
\label{sec:freezing_combinations}

We show results with different freezing combinations in Table \ref{tab:config-setup-averaged-dev-results}. We find that freezing the FFN component of the Decoder results in the biggest performance gains.

\subsection{Language-ID prediction on Cross-lingual Summarization}
We report the languages predicted by the Language Detection model from the Google Cloud Translation API\footnote{\url{https://cloud.google.com/translate/docs/basic/detecting-language}} \cite{caswell-etal-2020-language} for the XL-Sum$^{ar,en,ja,zh}$ task in Table~\ref{tab:lang_detection_1}. We find that \model achieves near perfect performance  for all target languages when freezing parts of the decoder (s7)--99\% of the text is generated in the correct target language--significantly outperforming all other model variants. Interestingly, \model hallucinates in the source language when the decoder is fine-tuned (s1), resulting in a drop down to only 2\% in the correct target language. \shared also benefits slightly from freezing parts of the decoder, with an improvement from 7\% to 18\% target language generation, however, this is no where close to the performance of \model.  

\subsection{Language-level Results}
\paragraph{XNLI.} 
We report XNLI validation results in Table~\ref{tab:xnli-dev-results} and test results in Table~\ref{tab:xnli-test-results}.
\paragraph{XQuAD.} We report XQuAD validation results in Table~\ref{tab:XQUAD_dev_results} and test results in Table~\ref{tab:Xquad_test_results}.
\paragraph{MASSIVE.}
 We report MASSIVE validation results in Table~\ref{tab:MASSIVE_DEV_RESULTS} and test results in Table~\ref{tab:MASSIVE_TEST_RESULTS}.
 \paragraph{TyDiQA.}
 We report TyDiQA  validation results in Table~\ref{tab:TYDIQA_dev_results}.

\paragraph{Multilingual XL-Sum}
 We report XL-Sum validation results in Tables~\ref{tab:XLSUM_dev_results_english_etc}, \ref{tab:XLSUM_dev_results_japanese_etc},\ref{tab:XLSUM_dev_results_portugues_etc}, \ref{tab:XLSUM_dev_results_somali_etc}, and \ref{tab:XLSUM_dev_results_ukrainian_etc} and test results in Table~\ref{tab:XLSUM_test_results}.
\paragraph{Zeroshot XL-Sum$^{en}$}
 We report XL-Sum validation results in Table~\ref{tab:XLSUM_en_dev_results} and test results in Table~\ref{tab:XLSUM_en_test_results}.
\paragraph{Zeroshot XL-Sum$^{ar,en,ja,zh}$}
 We report XL-Sum validation results in Table~\ref{tab:XLSUM_highres_dev_results} and test results in Table~\ref{tab:xlsum-arenjazh-test-results}.

\subsection{Language-level Pre-training Perplexities}
\begin{figure*}[t]
    \centering 
        \includegraphics[width=.99\linewidth]{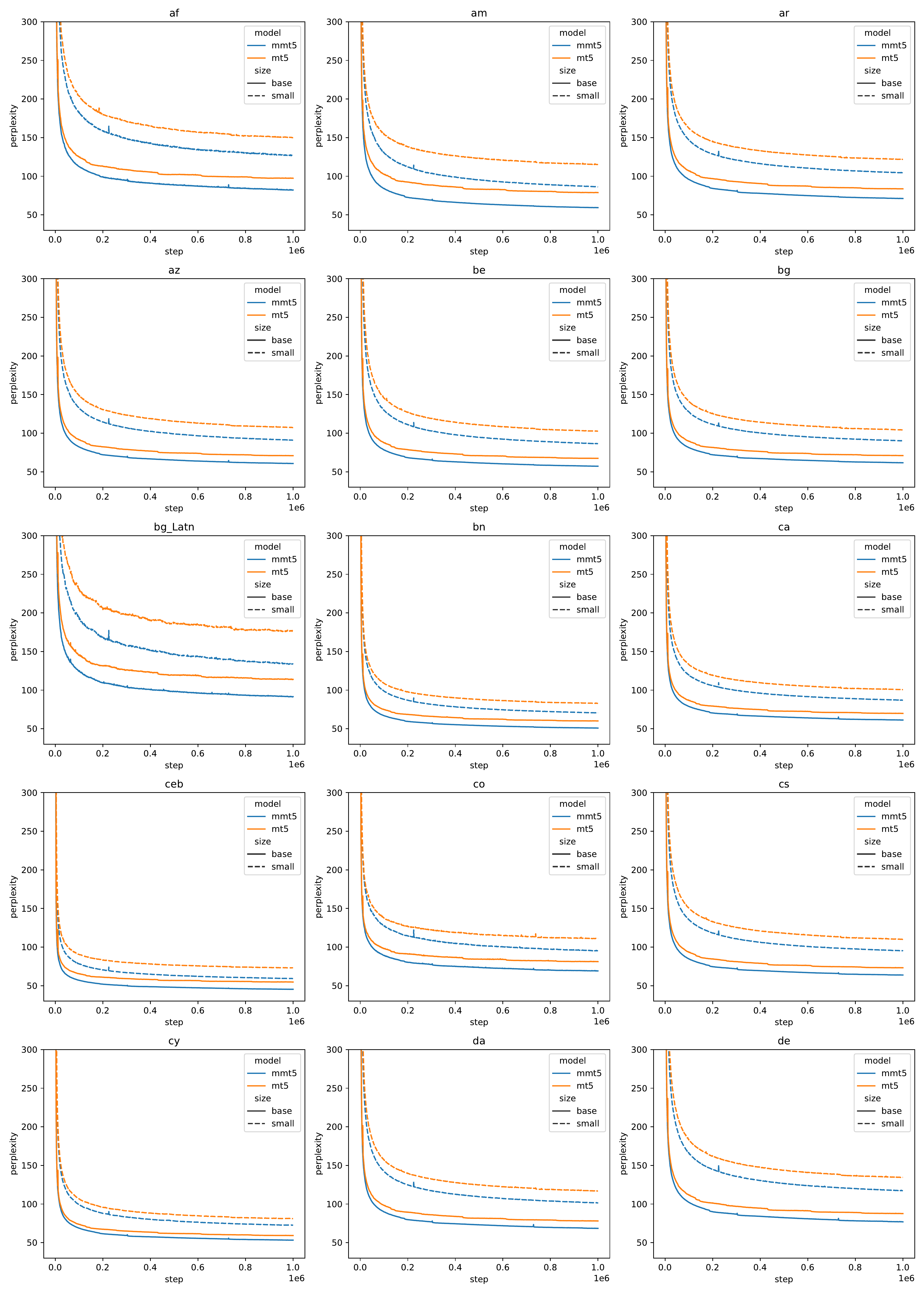}
    \caption{Per language perplexity of different model sizes for languages af-de. }
\label{fig:per_lang_perplexity_15}
\end{figure*}

\begin{figure*}[t]
    \centering 
        \includegraphics[width=.99\linewidth]{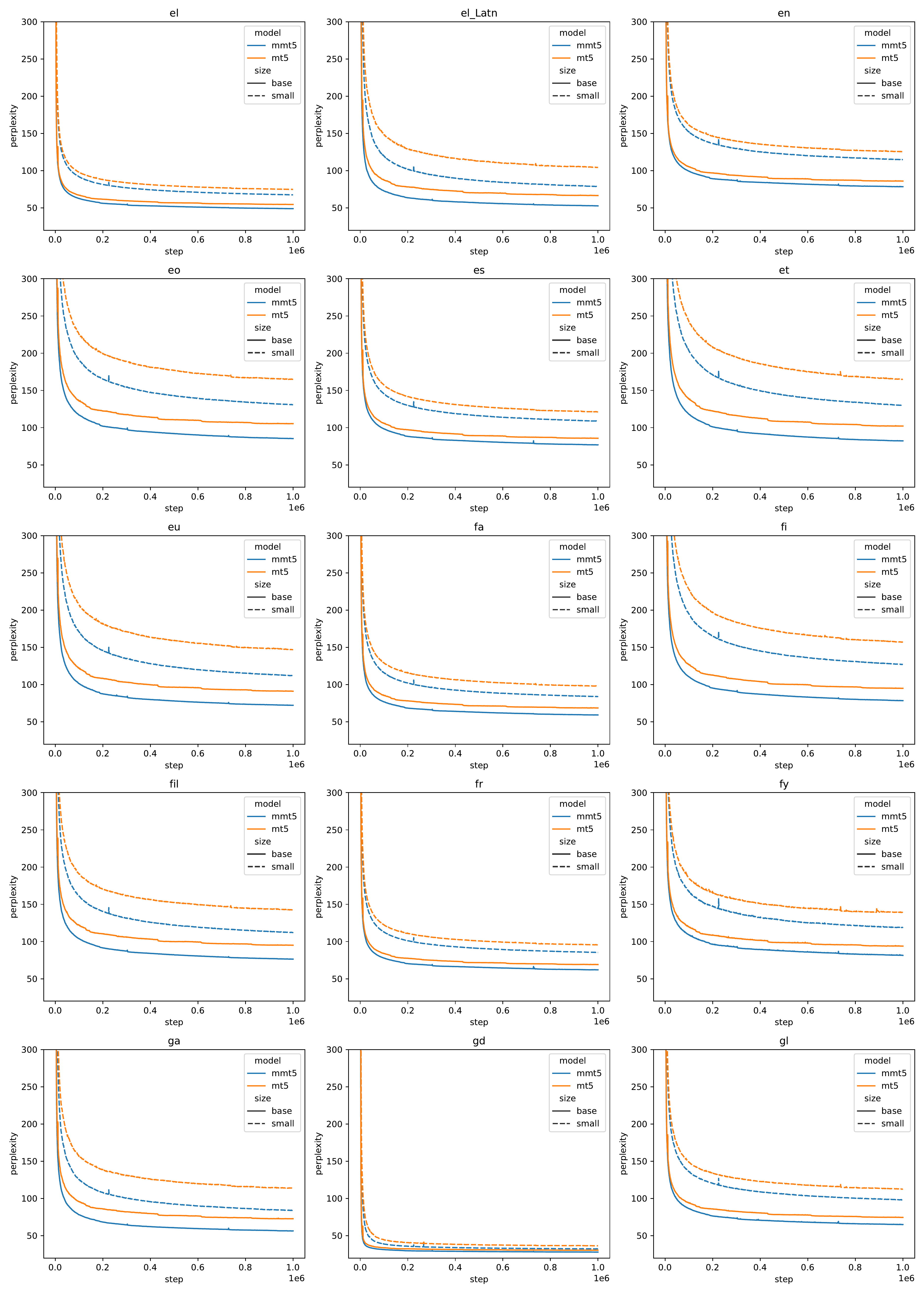}
    \caption{Per language perplexity of different model sizes  for languages el-gl.   }
\label{fig:per_lang_perplexity_30}
\end{figure*}

\begin{figure*}[t]
    \centering 
        \includegraphics[width=.99\linewidth]{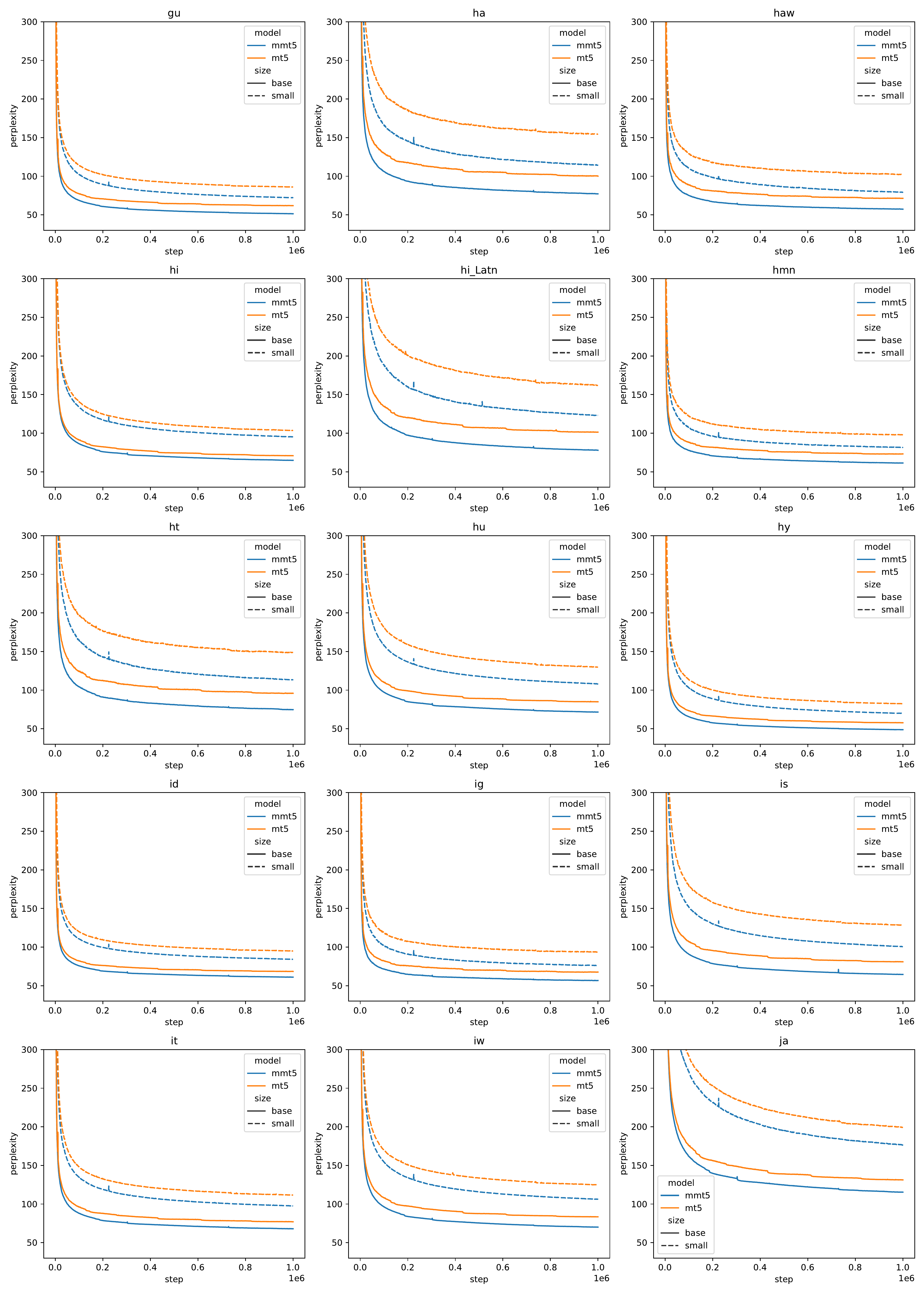}
    \caption{Per language perplexity of different model sizes  for languages gu-ja.   }
\label{fig:per_lang_perplexity_45}
\end{figure*}

\begin{figure*}[t]
    \centering 
        \includegraphics[width=.99\linewidth]{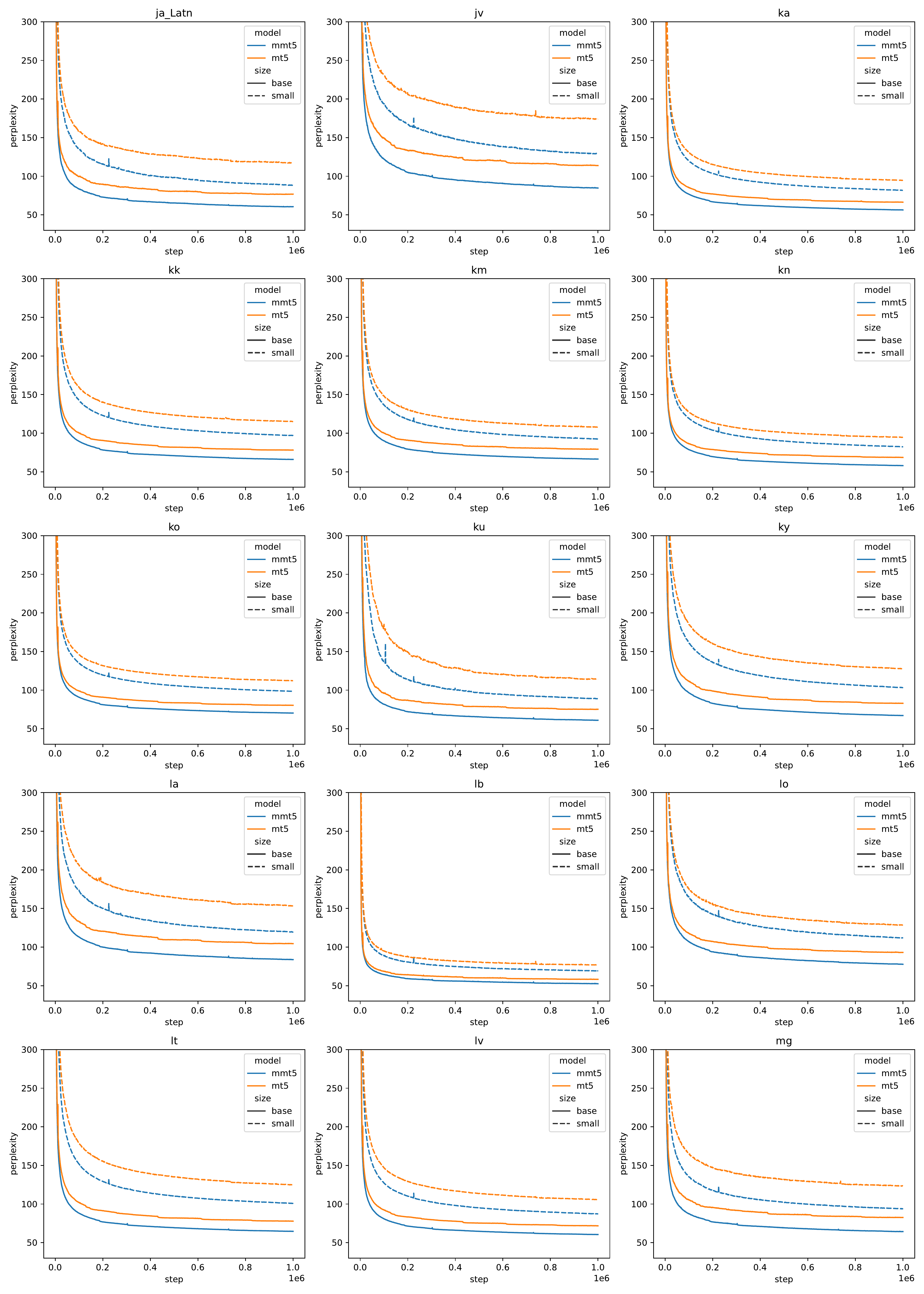}
    \caption{Per language perplexity of different model sizes for languages ja-mg.   }
\label{fig:per_lang_perplexity_60}
\end{figure*}

\begin{figure*}[t]
    \centering 
        \includegraphics[width=.99\linewidth]{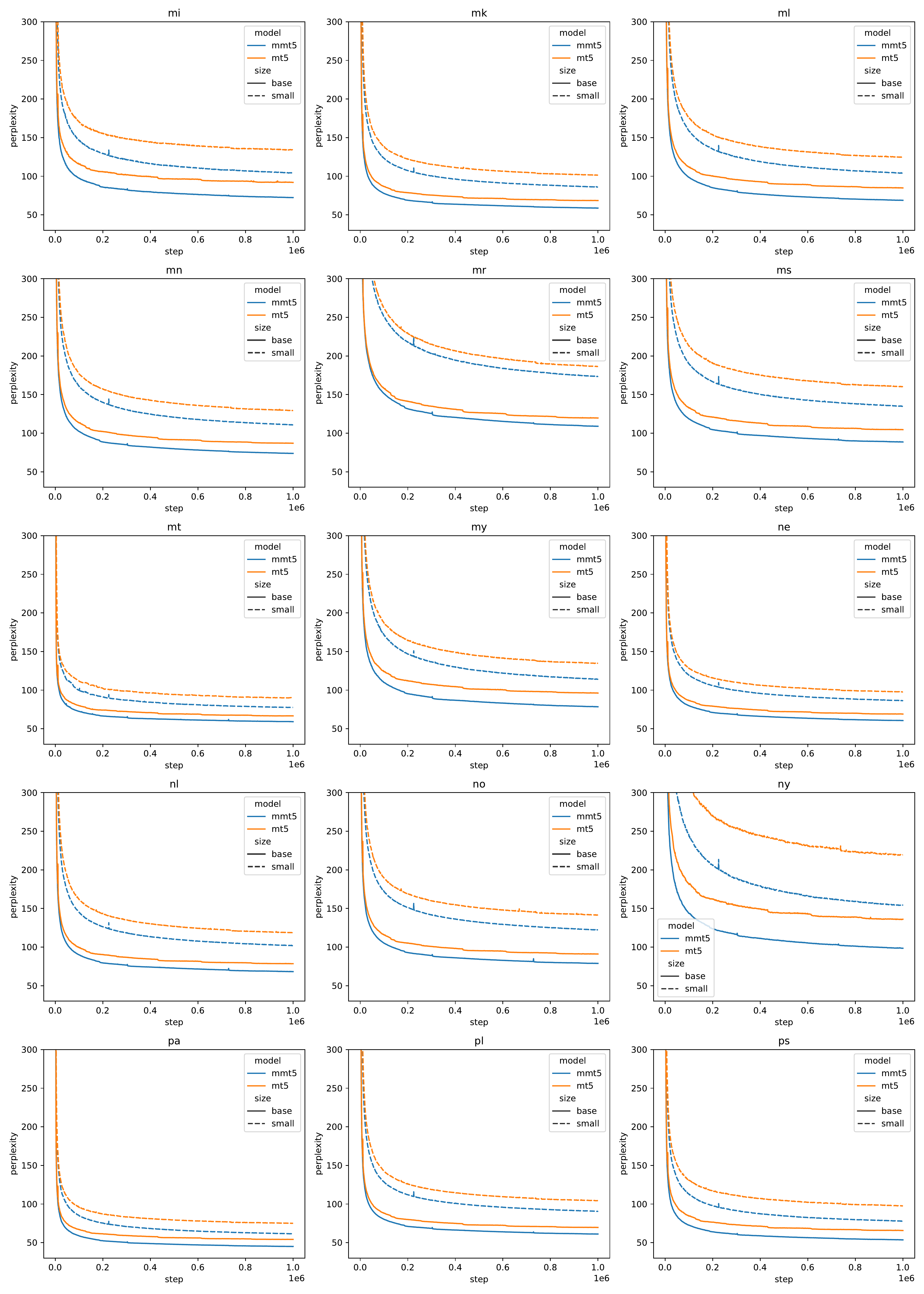}
    \caption{Per language perplexity of different model sizes  for languages mi-gs.   }
\label{fig:per_lang_perplexity_75}
\end{figure*}

\begin{figure*}[t]
    \centering 
        \includegraphics[width=.99\linewidth]{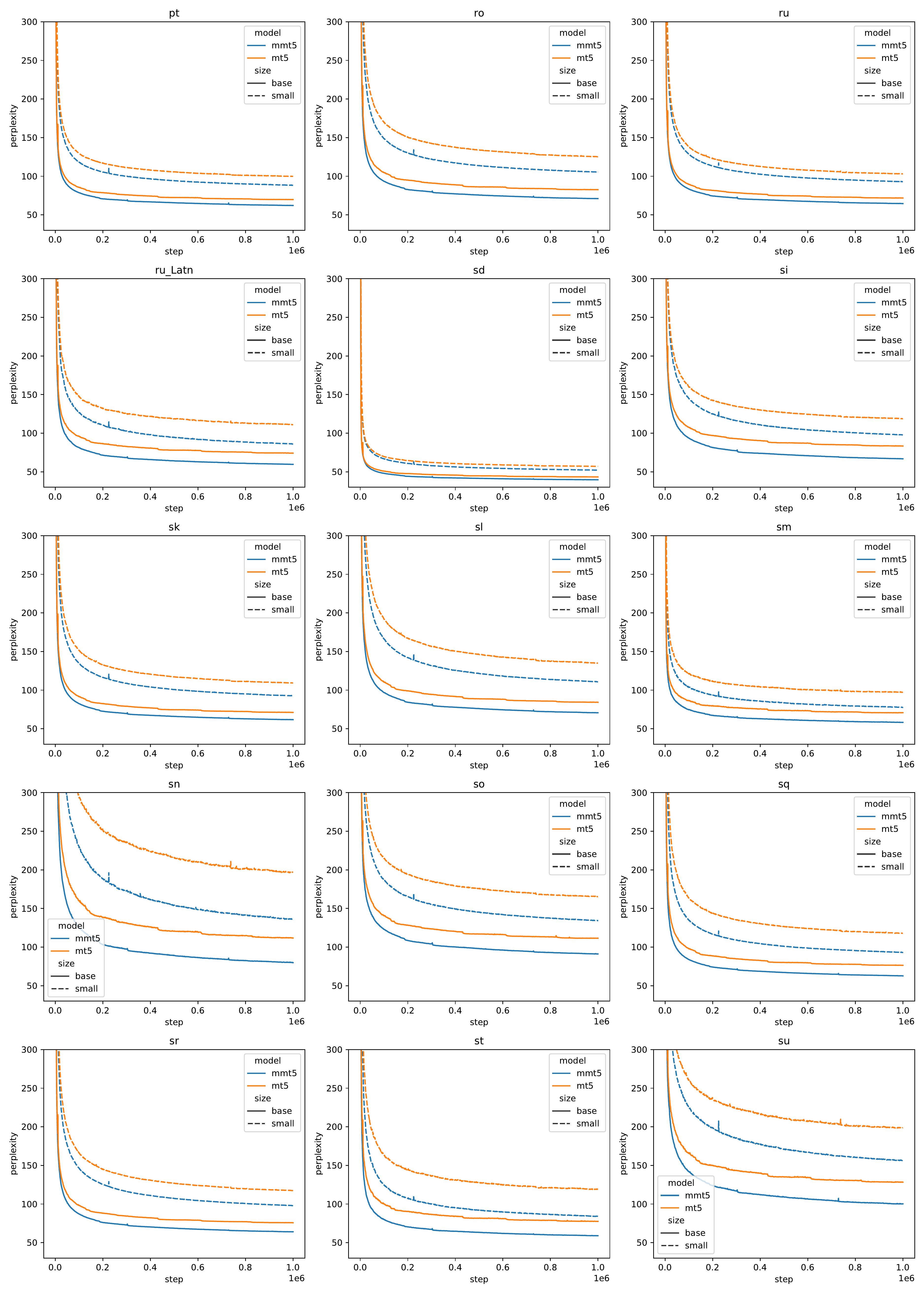}
    \caption{Per language perplexity of different model sizes  for languages pt-sg.   }
\label{fig:per_lang_perplexity_90}
\end{figure*}

\begin{figure*}[t]
    \centering 
        \includegraphics[width=.99\linewidth]{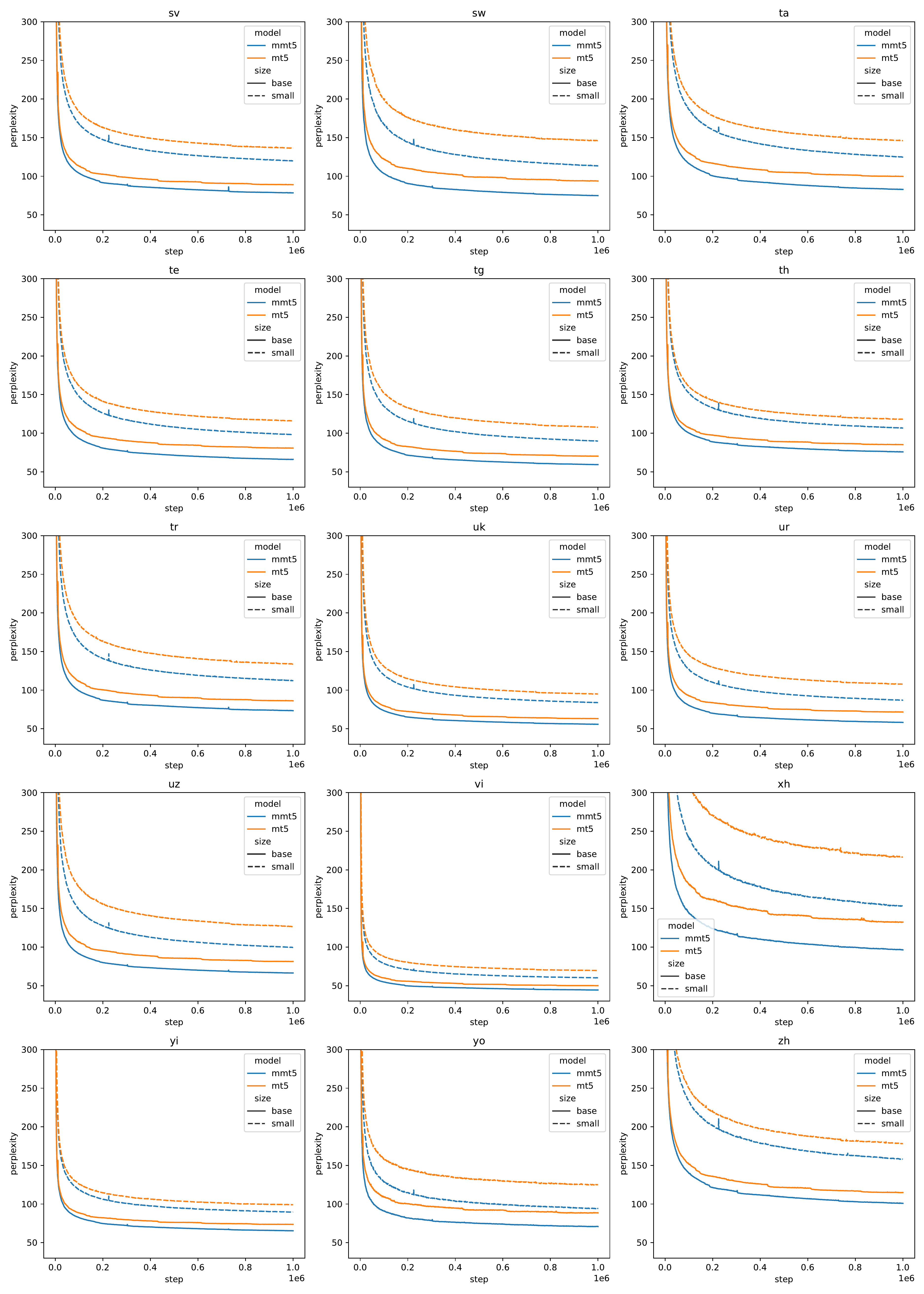}
    \caption{Per language perplexity of different model sizes  for languages sv-zh.   }
\label{fig:per_lang_perplexity_100}
\end{figure*}

We report the language-level perplexities of the different model variants and sizes in Figures~\ref{fig:per_lang_perplexity_15}, \ref{fig:per_lang_perplexity_30}, \ref{fig:per_lang_perplexity_45}, \ref{fig:per_lang_perplexity_60}, \ref{fig:per_lang_perplexity_75}, \ref{fig:per_lang_perplexity_90}, \ref{fig:per_lang_perplexity_100}.

\begin{table*}[]
\begin{center}
\def\arraystretch{0.87}
\resizebox{0.99\textwidth}{!}{%
\addtolength{\tabcolsep}{-0.5pt}
\begin{tabular}{l|c|c|cccc |cccccc|cc}
\toprule

&&&&&&&\multicolumn{6}{c}{Zero-Shot}  & \multicolumn{2}{|c}{Multi-Source}\\
&&&&&&&\multicolumn{2}{c}{XQuAD} &  \multicolumn{1}{c}{XNLI} &\multicolumn{1}{c}{XL-Sum$^{en}$} &\multicolumn{1}{c}{XL-Sum$^{ar,en,zh,ja}$} & \multicolumn{1}{c|}{MASSIVE}   & \multicolumn{1}{c}{XL-Sum} & \multicolumn{1}{c}{MASSIVE}\\
 &&&&&&&     \multicolumn{1}{c}{\textit{dev (en)}} &  
\multicolumn{1}{c}{\textit{test}}  &  \multicolumn{1}{c}{\textit{dev}}& \multicolumn{1}{c}{\textit{dev}} & \multicolumn{1}{c}{\textit{dev}} & \multicolumn{1}{c|}{\textit{dev}}  & \multicolumn{1}{c}{\textit{dev}} & \multicolumn{1}{c}{\textit{dev}}\\
cfg & Emb & Enc$_\text{LN}$  & Dec$_\text{LN}$ & Dec$_\text{Att}$ & Dec$_\text{CrossAtt}$ & Dec$_\text{FFN}$ &    \multicolumn{1}{c}{f1 / em} &  
\multicolumn{1}{c}{f1 / em}  &  \multicolumn{1}{c}{acc} & \multicolumn{1}{c}{Rg$_1$ / Rg$_2$ / Rg$_L$ }& \multicolumn{1}{c}{Rg$_1$ / Rg$_2$ / Rg$_L$ } & \multicolumn{1}{c|}{EM} & \multicolumn{1}{c}{Rg$_1$ / Rg$_2$ / Rg$_L$ }& \multicolumn{1}{c}{EM}\\
\midrule
    s1 &     &      &           &       &        &       & 90.7 /  83.6 & 66.9 /  49.3  &  75.5 & 15.4 / 2.0 / 14.0 & 18.7 /	6.1 /	16.8 & 32.1   & 41.2 / 22.4 / 32.4\\
    s2 &     &   \xmark&     \xmark&       &        &        & 90.7 /  83.4 & 65.6 /  48.0  &  75.0 & & & & \\
    s3 &  \xmark&      &        &       &        &        & 90.7 /  83.5 & 61.0 /  43.4  &  76.9 & & & & \\
    s4 &  \xmark&   \xmark&     \xmark&       &        &        & 90.9 /  83.6 & 64.6 /  47.1  &  77.5 & & & & \\
    s5 &  \xmark&      &     \xmark&    \xmark&     \xmark&    \xmark & 91.2 /  84.1 & 74.3 /  57.5  &  73.8 & & & & \\
    s6 &  \xmark&      &     \xmark&       &     \xmark&    \xmark & 91.9 /  85.1 & 75.8 /  59.5  &  75.6 & & & 43.2  & 41.2 / 22.4 / 32.6\\
    s7 &  \xmark&      &     \xmark&       &        &    \xmark & 91.8 /  85.1 & 75.8 /  59.8  &  77.3 & \textbf{19.7} /	\textbf{6.2} /	\textbf{16.4} & \textbf{34.7} /	\textbf{16.2} / \textbf{26.9} & 41.0   & \textbf{41.9} / \textbf{23.1} / \textbf{33.2}\\
    s8 &     &   \xmark&      \xmark&       &     \xmark&    \xmark    & 91.2 /  84.5 & 75.0 /  59.3 &    73.3 & & &  \\
    s9 &     &   \xmark&      \xmark&       &        &    \xmark    & 91.2 /  84.5 & 74.6 /  58.8  &   75.6 &  & &  \\
    s10 &  \xmark&   \xmark&      \xmark&       &     \xmark&    \xmark  & \textbf{92.1} /  \textbf{85.5} & \textbf{76.3} /  \textbf{60.3}  &  76.1  & & & \textbf{45.4}   & 40.8 / 22.1 / 32.3 & 66.78\\
   s11 &     &      &        \xmark&       &     \xmark&    \xmark  & 90.9 /  84.0 & 74.8 /  58.8  &  73.1 & & &  \\
   s12 &     &      &        \xmark&    \xmark&        &    \xmark  & 91.2 /  84.5 & 75.0 /  59.3  &  75.6 & & &  \\
   s13 &     &      &        \xmark&       &        &    \xmark  & 91.3 /  84.5 & 74.9 /  58.9  &  76.3 & & &  \\
   s14 &  \xmark&   \xmark&        \xmark&       &        &    \xmark  & 91.8 /  85.1 & 75.0 /  59.2  &  \textbf{77.7} & & & 39.9  & 41.8 / 23.0 / 33.1\\
\bottomrule
\end{tabular}
}
\end{center}
\caption{Results with different freezing combinations of \model base on different tasks. \xmark \,  indicates that the  component is frozen in the respective configuration Dev results for most. We always finetune the attention in the encoder (Enc$_{Att}$),  the feed forward layer in the encoder (Enc$_{FFN}$), and always freeze the modules in the encoder (Enc$_{Mod}$) and decoder (Dec$_{Mod}$). We find that configurations s1-s4 strongly underperform the respective other configurations (s5-s14), suggesting that freezing the feed forward layer of the decoder is essential for good cross-lingual transfer performance. 
}
\label{tab:config-setup-averaged-dev-results}
\end{table*}

\begin{table*}[]
\begin{center}
\def\arraystretch{0.87}
\resizebox{0.99\textwidth}{!}{%
\addtolength{\tabcolsep}{-0.5pt}
\begin{tabular}{cl rrrrrrrrrrrrrrr|r}
\toprule
   & model &       ar &    bg &    de &    el &    en &    es &    fr &    hi &    ru &    sw &    th &    tr &    ur &    vi &    zh & \textit{avg} \\
   \midrule
 small & \model &    \textbf{65.3} & \textbf{ 71.9} &  \textbf{70.2} &  \textbf{70.5} &  \textbf{81.8} &  \textbf{74.7} &  \textbf{73.4} &  \textbf{62.7} & \textbf{ 70.1} &  \textbf{63.8} &  \textbf{67.4} &  \textbf{64.2} &  \textbf{59.1} &  \textbf{66.7} &  \textbf{66.3} & \textbf{68.5} \\
 small & \shared &   63.8 &  69.1 &  67.6 &  68.1 &  80.0 &  71.6 &  69.3 &  60.4 &  68.7 &  53.2 &  64.1 &  59.1 &  58.4 &  63.9 &  64.4 & 65.5\\
 \midrule
 \midrule
 base & \model  &  \textbf{75.0} &  \textbf{81.2} &  \textbf{80.3} &  \textbf{79.5} &  \textbf{86.9} &  \textbf{82.6} &  \textbf{80.9} &  \textbf{73.4} &  \textbf{78.3} &  \textbf{74.0} &  \textbf{74.9} &  \textbf{76.3} &  \textbf{69.9} &  \textbf{76.7} &  \textbf{77.2} & \textbf{77.8} \\
 base & \shared &    72.7 &  78.7 &  77.6 &  77.3 &  85.5 &  80.7 &  78.5 &  70.7 &  77.7 &  66.6 &  73.0 &  72.8 &  67.7 &  73.8 &  73.5 & 75.1 \\

\bottomrule
\end{tabular}

}
\end{center}
\caption{ XNLI test results for all language. We select the checkpoint performing best on the validation set. 
}
\label{tab:xnli-test-results}
\end{table*}

\begin{table*}[]
\begin{center}
\def\arraystretch{0.87}
\resizebox{0.99\textwidth}{!}{%
\addtolength{\tabcolsep}{-0.5pt}
\begin{tabular}{llrrrrrrrrrrrrrrrrrrrrrrr}
\toprule
       & model &    ar &    de &    el &    en &    es &    hi &    ru &    th &    tr &    vi &    zh & \textit{avg} \\
 &  &     F1 \, /  EM  &     F1 \, /  EM  &     F1 \, /  EM  &     F1 \, /  EM  &     F1 \, /  EM  &     F1 \, /  EM  &     F1 \, /  EM  &     F1 \, /  EM  &     F1 \, /  EM  &     F1 \, /  EM  &     F1 \, /  EM &     F1 \, /  EM   \\ 
 \midrule
 \multirow{2}{*}{\rotatebox[origin=c]{90}{small}} & \model & \textbf{60.6} / \textbf{44.6} & 	\textbf{71.0} / \textbf{53.6} & 	\textbf{64.9} / \textbf{47.1} & 	\textbf{82.5} / \textbf{70.3} & 	\textbf{74.1} / \textbf{56.1} & 	\textbf{59.2} / \textbf{43.9} & 	\textbf{69.5} / \textbf{50.8} & 	\textbf{58.9} / \textbf{47.0} & 	\textbf{62.4} / \textbf{43.4} & 	\textbf{64.3} / \textbf{45.4} & 	64.2 / 52.4 & \textbf{66.5} /	\textbf{50.4} \\
  & \shared & 53.5 / 37.1 & 	67.2 / 48.7 & 	59.5 / 41.1 & 	81.7 / 69.7 & 	69.8 / 53.7 & 	54.7 / 40.8 & 	62.8 / 44.5 & 	50.3 / 37.9 & 	57.2 / 39.3 & 	59.7 / 40.9 & 	\textbf{64.7} / \textbf{54.0} & 61.9 /	46.2 \\
 \midrule
\midrule
 \multirow{2}{*}{\rotatebox[origin=c]{90}{base}} & \model & \textbf{74.2} / \textbf{57.6} & 	\textbf{79.5} / \textbf{63.0} & 	\textbf{77.6} / \textbf{59.9} & 	\textbf{86.7} / \textbf{74.5} & 	\textbf{79.2} / \textbf{61.3} & 	\textbf{72.4} / \textbf{56.1} & 	\textbf{77.6} / \textbf{58.7} & 	\textbf{69.3} / \textbf{59.3} & 	\textbf{74.5} / \textbf{55.9} & 	\textbf{74.2} / \textbf{54.2} & 	\textbf{74.4} / \textbf{63.1} & \textbf{76.3} /	\textbf{60.3} \\
 & \shared & 63.3 / 43.0 & 	75.9 / 57.2 & 	63.3 / 40.3 & 	84.3 / 71.9 & 	76.1 / 58.7 & 	62.8 / 47.1 & 	64.0 / 42.6 & 	59.6 / 48.5 & 	70.1 / 51.7 & 	70.4 / 50.4 & 	66.0 / 55.5 & 68.7 /	51.5 \\

\bottomrule
\end{tabular}
}
\end{center}
\caption{ XQuAD test set results for all languages. We select the checkpoint performing best on the English development set. 
}
\label{tab:Xquad_test_results}
\end{table*}

\begin{table}[]
\begin{center}
\tiny
\def\arraystretch{0.87}
\resizebox{0.3\textwidth}{!}{%
\addtolength{\tabcolsep}{-0.5pt}
\begin{tabular}{cc}
\toprule
Language & Exact Match (EM) \\
\midrule
\texttt{af\_ZA} & 57.5 \\
\texttt{am\_ET} & 29.6 \\
\texttt{ar\_SA} & 38.3 \\
\texttt{az\_AZ} & 41.3 \\
\texttt{bn\_BD} & 37.2 \\
\texttt{cy\_GB} & 35.5 \\
\texttt{da\_DK} & 60.5 \\
\texttt{de\_DE} & 55.3 \\
\texttt{el\_GR} & 49.6 \\
\texttt{en\_US} & 72.7 \\
\texttt{es\_ES} & 53.8 \\
\texttt{fa\_IR} & 48.2 \\
\texttt{fi\_FI} & 54.4 \\
\texttt{fr\_FR} & 51.4 \\
\texttt{hi\_IN} & 44.1 \\
\texttt{hu\_HU} & 47.4 \\
\texttt{hy\_AM} & 38.6 \\
\texttt{id\_ID} & 57.1 \\
\texttt{is\_IS} & 42.8 \\
\texttt{it\_IT} & 51.7 \\
\texttt{ja\_JP} & 42.5 \\
\texttt{jv\_ID} & 38.1 \\
\texttt{ka\_GE} & 38.9 \\
\texttt{km\_KH} & 40.7 \\
\texttt{kn\_IN} & 34.4 \\
\texttt{ko\_KR} & 39.1 \\
\texttt{lv\_LV} & 50.3 \\
\texttt{ml\_IN} & 36.0 \\
\texttt{mn\_MN} & 34.2 \\
\texttt{ms\_MY} & 52.5 \\
\texttt{my\_MM} & 33.8 \\
\texttt{nb\_NO} & 58.2 \\
\texttt{nl\_NL} & 57.5 \\
\texttt{pl\_PL} & 52.9 \\
\texttt{pt\_PT} & 56.0 \\
\texttt{ro\_RO} & 55.4 \\
\texttt{ru\_RU} & 50.9 \\
\texttt{sl\_SL} & 50.3 \\
\texttt{sq\_AL} & 48.3 \\
\texttt{sv\_SE} & 58.9 \\
\texttt{sw\_KE} & 43.0 \\
\texttt{ta\_IN} & 37.1 \\
\texttt{te\_IN} & 35.4 \\
\texttt{th\_TH} & 50.1 \\
\texttt{tr\_TR} & 47.9 \\
\texttt{ur\_PK} & 39.6 \\
\texttt{vi\_VN} & 44.9 \\
\texttt{zh\_CN} & 30.0 \\
\texttt{zh\_TW} & 28.2 \\
\midrule
Average & 46.0 \\
\bottomrule
\end{tabular}
}
\end{center}
\caption{MASSIVE Exact Match (EM) test accuracies of the best model (s10 modular) for all languages}
\label{tab:MASSIVE_TEST_RESULTS}
\end{table}

\begin{table}[]
\begin{center}
\def\arraystretch{0.87}
\resizebox{0.4\columnwidth}{!}{%
\addtolength{\tabcolsep}{-0.5pt}
\begin{tabular}{ccl rrrrrrrrrrrrrrrrrrrrrr}
\toprule
 & & &   en	 \\
 &  & cfg &    F1 \, /  EM  \\
 \midrule

  \multirow{10}{*}{\rotatebox[origin=c]{90}{Small}} & \multirow{5}{*}{\rotatebox[origin=c]{90}{\model}} & s1 &  85.6 /  77.6 \\
  & & s6 &  87.2 /  79.4 \\
& & s7 & \textbf{87.4} /  79.4 \\
& & s10 & 87.3 /  \textbf{79.6} \\
& & s14 & \textbf{87.4} /  79.2 \\
 \cmidrule{2-4}
 
& \multirow{5}{*}{\rotatebox[origin=c]{90}{\shared}} & s1 & 84.9 /  76.5 \\
& & s6 & 85.8 /  77.6 \\
&& s7 & 85.9 /  77.7 \\
&& s10 & 86.1 /  77.9 \\
&& s14 & 85.9 /  77.6 \\
 
 \midrule
\midrule
 \multirow{19}{*}{\rotatebox[origin=c]{90}{Base}} & \multirow{14}{*}{\rotatebox[origin=c]{90}{\model}} &   s1 & 90.7 /  83.6 \\
 & &  s2 & 90.7 /  83.4 \\
 & &    s3 & 90.7 /  83.5 \\
& &     s4 & 90.9 /  83.6 \\
    \cdashline{3-4}
&   &   s5 & 91.2 /  84.1 \\
& &     s6 &  91.9 /  85.1 \\
&   &   s7 & 91.8 /  85.1 \\
& &     s8 &  91.2 /  84.5 \\
&   &   s9 & 91.2 /  84.5 \\
& &        s10 & \textbf{92.1} /  \textbf{85.5} \\
 &   & s11 & 90.9 /  84.0 \\
 &   & s12 &  91.2 /  84.5 \\
&    & s13 &91.3 /  84.5 \\
&  &   s14 & 91.8 /  85.1 \\
   \cmidrule{2-4}
&  \multirow{5}{*}{\rotatebox[origin=c]{90}{\shared.}} &  s1 & 89.9 /  82.5 \\
&& s6 & 90.2  / 83.0 \\
& & s7 &90.5 /  83.6 \\

& & s10 &90.2 /  82.8 \\

& & s14 & 90.4 /  83.5 \\

\bottomrule
\end{tabular}
}
\end{center}
\caption{  XQuAD validation results for English across the different freezing configurations. 
}
\label{tab:XQUAD_dev_results}
\end{table}

\begin{table*}[]
\begin{center}
\def\arraystretch{0.87}
\resizebox{0.96\textwidth}{!}{%
\addtolength{\tabcolsep}{-0.5pt}
\begin{tabular}{ll|rrrrr|rrrrr|rrrrr|rrrrr|rrrrr}
\toprule
  tgt lang && \multicolumn{5}{c}{am} & \multicolumn{5}{c}{ar} & \multicolumn{5}{c}{az} & \multicolumn{5}{c}{bn} & \multicolumn{5}{c}{cy} \\
            pred  lang  & cfg &   \textit{am} &    ar &    en &    ja &    zh &  \textit{ar} &   ar &   en &   ja &   zh &   \textit{az} &    ar &    en &    ja &    zh &   \textit{bn} &   ar &    en &    ja &    zh &   \textit{cy} &    ar &    en &   ja &    zh \\
 
\midrule
  \model &s1 &  0.00 &  0.50 &  0.07 &  0.02 &  0.36 &  \textbf{1.0} &  1.0 &  0.0 &  0.0 &  0.0 &  0.00 &  0.18 &  0.29 &  0.06 &  0.29 &  0.00 &  0.4 &  0.10 &  0.05 &  0.41 &  0.06 &  0.05 &  0.84 &  0.0 &  0.01 \\
  \model &s7 &  \textbf{0.99} &  0.00 &  0.00 &  0.00 &  0.00 &  \textbf{1.0} &  1.0 &  0.0 &  0.0 &  0.0 &  \textbf{1.00} &  0.00 &  0.00 &  0.00 &  0.00 &  \textbf{1.00} &  0.0 &  0.00 &  0.00 &  0.00 &  \textbf{0.91} &  0.00 &  0.08 &  0.0 &  0.00 \\
  \shared &s1 &  0.03 &  0.96 &  0.00 &  0.00 &  0.01 &  \textbf{1.0} &  1.0 &  0.0 &  0.0 &  0.0 &  0.08 &  0.12 &  0.31 &  0.27 &  0.09 &  0.02 &  0.3 &  0.02 &  0.43 &  0.19 &  0.13 &  0.10 &  0.76 &  0.0 &  0.00 \\
 \shared  & s7 &  0.02 &  0.95 &  0.00 &  0.00 &  0.00 &  \textbf{1.0} &  1.0 &  0.0 &  0.0 &  0.0 &  0.36 &  0.03 &  0.35 &  0.05 &  0.02 &  0.13 &  0.1 &  0.13 &  0.47 &  0.11 &  0.22 &  0.01 &  0.76 &  0.0 &  0.00 \\

\midrule
 \midrule
 
           tgt lang & &  \multicolumn{5}{c}{en} & \multicolumn{5}{c}{es} & \multicolumn{5}{c}{fa} & \multicolumn{5}{c}{fr} & \multicolumn{5}{c}{gd} \\
             pred  lang &  cfg &   \textit{en} &   ar &   en &   ja &   zh &   \textit{es} &    ar &    en &    ja &    zh &   \textit{fa} &    ar &    en &   ja &   zh &   \textit{fr} &    ar &    en &   ja &    zh &   \textit{gd} &    ar &    en &   ja &    zh \\
 
\midrule
  \model  & s1 &  \textbf{1.0} &  0.0 &  1.0 &  0.0 &  0.0 &  0.03 &  0.03 &  0.73 &  0.01 &  0.07 &  0.02 &  0.96 &  0.01 &  0.0 &  0.0 &  0.05 &  0.07 &  0.74 &  0.0 &  0.02 &  0.25 &  0.01 &  0.67 &  0.0 &  0.01 \\
  \model  & s7 &  \textbf{1.0} &  0.0 &  1.0 &  0.0 &  0.0 &  \textbf{0.99} &  0.00 &  0.00 &  0.00 &  0.00 &  \textbf{1.00} &  0.00 &  0.00 &  0.0 &  0.0 &  \textbf{1.00} &  0.00 &  0.00 &  0.0 &  0.00 &  \textbf{1.00} &  0.00 &  0.00 &  0.0 &  0.00 \\
 \shared  & s1 &  \textbf{1.0} &  0.0 &  1.0 &  0.0 &  0.0 &  0.02 &  0.00 &  0.95 &  0.00 &  0.00 &  0.01 &  0.99 &  0.00 &  0.0 &  0.0 &  0.01 &  0.00 &  0.98 &  0.0 &  0.00 &  0.23 &  0.01 &  0.76 &  0.0 &  0.00 \\
  \shared  & s7 &  \textbf{1.0} &  0.0 &  1.0 &  0.0 &  0.0 &  0.05 &  0.00 &  0.91 &  0.00 &  0.00 &  0.13 &  0.86 &  0.00 &  0.0 &  0.0 &  0.04 &  0.00 &  0.94 &  0.0 &  0.00 &  0.56 &  0.00 &  0.43 &  0.0 &  0.00 \\
 
 \midrule
\midrule

          tgt lang  & & \multicolumn{5}{c}{gu} & \multicolumn{5}{c}{ha} & \multicolumn{5}{c}{hi} & \multicolumn{5}{c}{id} & \multicolumn{5}{c}{ig} \\
          pred  lang  &  cfg &   \textit{gu} &    ar &    en &    ja &    zh &   \textit{ha} &    ar &    en &    ja &    zh &   \textit{hi} &    ar &    en &    ja &    zh &   \textit{id} &    ar &    en &    ja &    zh &   \textit{ig} &    ar &    en &    ja &    zh \\
 
\midrule
  \model  & s1 &  0.00 &  0.24 &  0.18 &  0.10 &  0.41 &  0.03 &  0.17 &  0.62 &  0.01 &  0.01 &  0.00 &  0.33 &  0.26 &  0.04 &  0.30 &  0.01 &  0.13 &  0.62 &  0.03 &  0.13 &  0.10 &  0.25 &  0.33 &  0.02 &  0.17 \\
  \model  & s7 &  \textbf{1.00} &  0.00 &  0.00 &  0.00 &  0.00 &  \textbf{1.00} &  0.00 &  0.00 &  0.00 &  0.00 & \textbf{ 1.00} &  0.00 &  0.00 &  0.00 &  0.00 &  \textbf{0.98} &  0.00 &  0.00 &  0.00 &  0.00 &  \textbf{1.00} &  0.00 &  0.00 &  0.00 &  0.00 \\
 \shared  & s1 &  0.13 &  0.16 &  0.01 &  0.56 &  0.13 &  0.11 &  0.03 &  0.83 &  0.00 &  0.00 &  0.03 &  0.23 &  0.02 &  0.64 &  0.05 &  0.07 &  0.14 &  0.73 &  0.01 &  0.02 &  0.63 &  0.01 &  0.34 &  0.00 &  0.00 \\
  \shared  & s7 &  0.23 &  0.05 &  0.12 &  0.49 &  0.06 &  0.18 &  0.01 &  0.76 &  0.00 &  0.00 &  0.12 &  0.08 &  0.23 &  0.48 &  0.01 &  0.25 &  0.03 &  0.67 &  0.00 &  0.00 &  0.66 &  0.00 &  0.31 &  0.00 &  0.00 \\

 \midrule
\midrule

  tgt lang & &  \multicolumn{5}{c}{ja} & \multicolumn{5}{c}{ko} & \multicolumn{5}{c}{ky} & \multicolumn{5}{c}{mr} & \multicolumn{5}{c}{my} \\
     pred  lang &  cfg &   \textit{ja} &   ar &   en &   ja &   zh &   \textit{ko} &    ar &    en &    ja &    zh &   \textit{ky} &    ar &    en &    ja &    zh &   \textit{mr} &    ar &    en &    ja &    zh &   \textit{my} &    ar &    en &    ja &    zh \\
 
\midrule
  \model  & s1 &  \textbf{1.0} &  0.0 &  0.0 &  1.0 &  0.0 &  0.00 &  0.01 &  0.04 &  0.50 &  0.43 &  0.00 &  0.05 &  0.12 &  0.13 &  0.65 &  0.00 &  0.12 &  0.24 &  0.04 &  0.55 &  0.00 &  0.24 &  0.03 &  0.30 &  0.27 \\
  \model  & s7 &  \textbf{1.0} &  0.0 &  0.0 &  1.0 &  0.0 &  \textbf{1.00} &  0.00 &  0.00 &  0.00 &  0.00 &  \textbf{0.99} &  0.00 &  0.00 &  0.00 &  0.00 &  \textbf{0.96} &  0.00 &  0.00 &  0.00 &  0.00 &  \textbf{1.00} &  0.00 &  0.00 &  0.00 &  0.00 \\
  \shared  & s1 &  \textbf{1.0} &  0.0 &  0.0 &  1.0 &  0.0 &  0.01 &  0.00 &  0.00 &  0.99 &  0.00 &  0.02 &  0.15 &  0.04 &  0.56 &  0.14 &  0.07 &  0.09 &  0.05 &  0.64 &  0.08 &  0.05 &  0.03 &  0.00 &  0.91 &  0.01 \\
  \shared  & s7 &  \textbf{1.0} &  0.0 &  0.0 &  1.0 &  0.0 &  0.02 &  0.00 &  0.00 &  0.98 &  0.00 &  0.08 &  0.02 &  0.24 &  0.40 &  0.07 &  0.11 &  0.02 &  0.19 &  0.37 &  0.08 &  0.20 &  0.00 &  0.00 &  0.78 &  0.01 \\

 \midrule
\midrule

  tgt lang  & & \multicolumn{5}{c}{ne} & \multicolumn{5}{c}{pa} & \multicolumn{5}{c}{ps} & \multicolumn{5}{c}{pt} & \multicolumn{5}{c}{ru} \\
  pred  lang  & cfg  &   \textit{ne} &    ar &    en &    ja &    zh &   \textit{pa} &    ar &    en &    ja &    zh &   \textit{ps} &    ar &    en &   ja &   zh &   \textit{pt} &    ar &    en &    ja &    zh &   \textit{ru} &    ar &    en &    ja &    zh \\
 
\midrule
  \model  & s1 &  0.00 &  0.25 &  0.10 &  0.10 &  0.47 &  0.01 &  0.22 &  0.22 &  0.07 &  0.42 &  0.03 &  0.92 &  0.00 &  0.0 &  0.0 &  0.04 &  0.08 &  0.74 &  0.01 &  0.03 &  0.00 &  0.50 &  0.21 &  0.02 &  0.20 \\
  \model  & s7 &  \textbf{0.99} &  0.00 &  0.00 &  0.00 &  0.00 &  \textbf{1.00} &  0.00 &  0.00 &  0.00 &  0.00 &  \textbf{1.00} &  0.00 &  0.00 &  0.0 &  0.0 &  \textbf{1.00} &  0.00 &  0.00 &  0.00 &  0.00 &  \textbf{1.00} &  0.00 &  0.00 &  0.00 &  0.00 \\
  \shared  & s1 &  0.00 &  0.15 &  0.00 &  0.67 &  0.14 &  0.05 &  0.24 &  0.04 &  0.48 &  0.16 &  0.02 &  0.96 &  0.00 &  0.0 &  0.0 &  0.02 &  0.00 &  0.97 &  0.00 &  0.00 &  0.07 &  0.53 &  0.28 &  0.01 &  0.04 \\
  \shared  & s7 &  0.01 &  0.02 &  0.03 &  0.82 &  0.09 &  0.18 &  0.12 &  0.19 &  0.40 &  0.03 &  0.12 &  0.82 &  0.02 &  0.0 &  0.0 &  0.07 &  0.00 &  0.91 &  0.00 &  0.00 &  0.27 &  0.05 &  0.54 &  0.00 &  0.00 \\

 \midrule
\midrule

  tgt lang &  & \multicolumn{5}{c}{si} & \multicolumn{5}{c}{so} & \multicolumn{5}{c}{sr} & \multicolumn{5}{c}{sw} & \multicolumn{5}{c}{ta} \\
 pred  lang  & cfg  &   \textit{si} &    ar &    en &    ja &    zh &   \textit{so} &    ar &    en &    ja &    zh &   \textit{sr} &     ar &     en &     ja &     zh &   \textit{sw} &    ar &    en &    ja &    zh &   \textit{ta} &    ar &    en &    ja &    zh \\ 
\midrule
  \model  & s1 &  0.00 &  0.30 &  0.22 &  0.11 &  0.26 &  0.01 &  0.22 &  0.45 &  0.03 &  0.04 &  0.00 &  0.110 &  0.590 &  0.035 &  0.185 &  0.01 &  0.18 &  0.51 &  0.02 &  0.06 &  0.01 &  0.24 &  0.16 &  0.14 &  0.33 \\
  \model  & s7 &  \textbf{1.00} &  0.00 &  0.00 &  0.00 &  0.00 &  \textbf{1.00} &  0.00 &  0.00 &  0.00 &  0.00 &  \textbf{0.95} &  0.000 &  0.010 &  0.000 &  0.000 &  \textbf{1.00} &  0.00 &  0.00 &  0.00 &  0.00 &  \textbf{1.00} &  0.00 &  0.00 &  0.00 &  0.00 \\
 \shared  & s1 &  0.02 &  0.48 &  0.01 &  0.40 &  0.02 &  0.04 &  0.65 &  0.25 &  0.00 &  0.00 &  0.01 &  0.375 &  0.525 &  0.005 &  0.005 &  0.01 &  0.84 &  0.13 &  0.00 &  0.00 &  0.07 &  0.09 &  0.02 &  0.68 &  0.07 \\
  \shared  & s7 &  0.07 &  0.63 &  0.10 &  0.11 &  0.00 &  0.06 &  0.71 &  0.15 &  0.00 &  0.00 &  0.02 &  0.260 &  0.545 &  0.000 &  0.000 &  0.02 &  0.87 &  0.09 &  0.00 &  0.00 &  0.32 &  0.05 &  0.06 &  0.46 &  0.03 \\

 \midrule
\midrule

  tgt lang & &  \multicolumn{5}{c}{te} & \multicolumn{5}{c}{th} & \multicolumn{5}{c}{tr} & \multicolumn{5}{c}{uk} & \multicolumn{5}{c}{ur} \\
  pred  lang  & cfg  &   \textit{te} &    ar &    en &    ja &    zh &   \textit{th} &    ar &    en &    ja &    zh &   \textit{tr} &    ar &    en &    ja &    zh &   \textit{uk} &    ar &    en &    ja &    zh &  \textit{ur} &    ar &    en &    ja &    zh \\
 
\midrule
  \model  & s1 &  0.01 &  0.35 &  0.14 &  0.10 &  0.32 &  0.00 &  0.13 &  0.04 &  0.01 &  0.79 &  0.01 &  0.04 &  0.49 &  0.13 &  0.18 &  0.00 &  0.24 &  0.32 &  0.04 &  0.32 &  0.01 &  0.88 &  0.04 &  0.00 &  0.00 \\
 \model  & s7 &  \textbf{1.00} &  0.00 &  0.00 &  0.00 &  0.00 &  \textbf{1.00} &  0.00 &  0.00 &  0.00 &  0.00 &  \textbf{1.00} &  0.00 &  0.00 &  0.00 &  0.00 &  \textbf{1.00} &  0.00 &  0.00 &  0.00 &  0.00 &  \textbf{1.00} &  0.00 &  0.00 &  0.00 &  0.00 \\
  \shared  & s1 &  0.09 &  0.30 &  0.02 &  0.49 &  0.09 &  0.08 &  0.18 &  0.01 &  0.00 &  0.68 &  0.08 &  0.01 &  0.44 &  0.39 &  0.03 &  0.02 &  0.79 &  0.12 &  0.00 &  0.03 &  0.02 &  0.56 &  0.01 &  0.20 &  0.14 \\
  \shared  & s7 &  0.31 &  0.28 &  0.06 &  0.22 &  0.05 &  0.29 &  0.00 &  0.00 &  0.04 &  0.60 &  0.37 &  0.00 &  0.27 &  0.24 &  0.00 &  0.18 &  0.41 &  0.31 &  0.00 &  0.00 &  0.20 &  0.20 &  0.06 &  0.24 &  0.03 \\

 \midrule
\midrule

     tgt lang &  & \multicolumn{5}{c}{uz} & \multicolumn{5}{c}{vi} & \multicolumn{5}{c}{yo} & \multicolumn{5}{c}{zh}  & \multicolumn{5}{c}{\textit{avg  (excluding source langs)}} \\
      pred  lang &  cfg &    \textit{uz} &    ar &    en &    ja &    zh &   \textit{vi} &    ar &    en &    ja &    zh &   \textit{yo} &    ar &    en &    ja &    zh &  \textit{zh} &   ar &   en &   ja &   zh &   \textit{tgt} & ar &   en &   ja &   zh \\ 
\midrule
  \model  & s1 &  0.01 &  0.14 &  0.15 &  0.14 &  0.31 &  0.02 &  0.26 &  0.28 &  0.03 &  0.31 &  0.02 &  0.25 &  0.29 &  0.02 &  0.17 &  \textbf{1.0} &  0.0 &  0.0 &  0.0 &  1.0 & 0.02 & 0.25	& 0.32 &	0.07 &		0.24\\
  \model  & s7 &  \textbf{0.99} &  0.00 &  0.00 &  0.00 &  0.00 &  \textbf{1.00} &  0.00 &  0.00 &  0.00 &  0.00 & \textbf{ 1.00} &  0.00 &  0.00 &  0.00 &  0.00 &  \textbf{1.0} &  0.0 &  0.0 &  0.0 &  1.0 & \textbf{0.99} & 0.00	& 0.00 & 	0.00 &	0.00\\
 \shared  & s1 &  0.00 &  0.36 &  0.01 &  0.45 &  0.11 &  0.08 &  0.24 &  0.49 &  0.00 &  0.16 &  0.22 &  0.07 &  0.66 &  0.00 &  0.01 &  \textbf{1.0} &  0.0 &  0.0 &  0.0 &  1.0 & 0.07 & 0.29 &	0.29 &	0.24   & 0.07	 \\
 \shared  & s7 &  0.02 &  0.08 &  0.19 &  0.46 &  0.03 &  0.31 &  0.04 &  0.61 &  0.00 &  0.00 &  0.44 &  0.00 &  0.50 &  0.00 &  0.00 &  \textbf{1.0} &  0.0 &  0.0 &  0.0 &  1.0 & 0.18 & 0.19 &	0.31 &	0.19 &		0.03\\

\bottomrule
\end{tabular}
}
\end{center}
\caption{Language prediction results on the XL-Sum$^{ar,en,ja,zh}$ task setup. The generated summarization text is passed into the language prediction model. We report the percentage of text which the model predicts to be in the correct \textit{target} language, as well as each of the 4 source languages. It is possible that another language was predicted, the numbers therefore do not need to sum up to 1.0. }
\label{tab:lang_detection_1}
\end{table*}

\begin{table*}[]
\begin{center}
\def\arraystretch{0.87}
\resizebox{0.99\textwidth}{!}{%
\addtolength{\tabcolsep}{-0.5pt}
\begin{tabular}{ccl rrrrrrrrrrrrrrr|r}
\toprule
 &  & cfg &       ar &    bg &    de &    el &    en &    es &    fr &    hi &    ru &    sw &    th &    tr &    ur &    vi &    zh & \textit{avg} \\

\midrule
\multirow{6}{*}{\rotatebox[origin=c]{90}{Small}}  & \multirow{5}{*}{\rotatebox[origin=c]{90}{\model}} 
& s1 &     63.7 &  68.4 &  68.4 &  66.7 &  77.6 &  71.6 &  69.3 &  60.8 &  66.3 &  60.2 &  62.2 &  62.5 &  56.2 &  64.5 &  63.4 & 65.4 \\
& & s6 &     63.5 &  69.2 &  69.2 &  68.2 &  \textbf{81.7} &  73.9 &  71.7 &  62.2 &  67.1 &  60.8 &  64.8 &  62.3 &  54.7 &  61.3 &  64.2  & 66.3\\
   & & s7 &     \textbf{65.4} &  \textbf{70.8} &  \textbf{70.5} &  \textbf{70.1} &  81.5 &  \textbf{74.5} &  \textbf{73.0} &  \textbf{63.1} &  \textbf{69.1} &  \textbf{62.7} &  \textbf{66.1} &  \textbf{63.1} &  \textbf{58.4} &  \textbf{65.7} &  65.7 & \textbf{68.0} \\
  &   & s10 &     63.9 &  69.6 &  68.9 &  69.4 &  80.6 &  73.7 &  70.4 &  62.4 &  67.8 &  59.9 &  65.2 &  61.4 &  54.6 &  62.6 &  65.6 & 66.4\\
     &     & s14 &     64.7 &  70.7 &  70.4 &  69.5 &  81.6 &  74.5 &  72.6 &  63.2 &  68.8 &  60.9 &  65.3 &  62.4 &  58.0 &  65.1 &  \textbf{65.9} & 67.6 \\
      \cline{2-19}
 & \multirow{5}{*}{\rotatebox[origin=c]{90}{\shared}} & s1 &     63.2 &  68.6 &  67.8 &  67.3 &  80.2 &  70.8 &  70.1 &  59.8 &  67.6 &  52.9 &  62.5 &  58.2 &  56.9 &  63.5 &  63.2  & 64.8\\
   &  &   s6 &     58.8 &  64.5 &  64.2 &  62.4 &  77.2 &  67.3 &  65.9 &  55.0 &  64.0 &  48.2 &  60.3 &  53.1 &  50.9 &  58.6 &  61.2 & 60.8 \\
\bottomrule
\end{tabular}

}
\end{center}
\caption{ XNLI validation results  for all languages. We report the results of different freezing configurations.
}
\label{tab:xnli-dev-results}
\end{table*}

\begin{table*}[]
\begin{center}
\def\arraystretch{0.87}
\resizebox{0.99\textwidth}{!}{%
\addtolength{\tabcolsep}{-0.5pt}
\begin{tabular}{llrrrrrrr}
\toprule
       & lang &    amharic & arabic & azerbaijani & bengali & burmese & chinese\_simplified & chinese\_traditional \\
        &  &    Rg$_1$ / Rg$_2$ / Rg$_L$ &   Rg$_1$ / Rg$_2$ / Rg$_L$  &   Rg$_1$ / Rg$_2$ / Rg$_L$  &   Rg$_1$ / Rg$_2$ / Rg$_L$  &   Rg$_1$ / Rg$_2$ / Rg$_L$  &   Rg$_1$ / Rg$_2$ / Rg$_L$  &   Rg$_1$ / Rg$_2$ / Rg$_L$  \\
\midrule
\multirow{2}{*}{small} & \model &  29.7 / 13.4 / 22.7 & 	37.5 / 17.3 / 29.4 & 	18.5 / 6.8 / 15.5 & 	27.6 / 13.1 / 22.5 & 	38.7 / 18.2 / 30.0 & 	38.7 / 19.4 / 33.3 & 	39.3 / 20.9 / 33.6 \\
 & \shared & 13.2 / 3.4 / 12.2 & 	37.5 / 17.4 / 29.4 & 	13.3 / 2.5 / 12.1 & 	7.6 / 1.1 / 7.4 & 	12.7 / 2.0 / 12.1 & 	38.2 / 19.1 / 33.1 & 	39.1 / 20.9 / 33.4 \\
\midrule
\midrule
\multirow{2}{*}{base} & \model &  34.2 / 17.1 / 25.8 & 	42.3 / 22.2 / 33.4 & 	26.4 / 11.1 / 21.3 & 	33.8 / 17.0 / 26.1 & 	40.0 / 18.9 / 30.7 & 	48.1 / 29.1 / 42.9 & 	49.0 / 30.8 / 43.1    \\
 & \shared &   15.0 / 4.5 / 13.7 & 	40.2 / 20.1 / 31.5 & 	16.0 / 4.4 / 14.2 & 	11.9 / 3.4 / 11.2 & 	15.7 / 3.7 / 14.5 & 	44.2 / 26.1 / 39.2 & 	45.2 / 27.7 / 39.4    \\
\bottomrule

\\

\toprule
       & lang &     english & french & gujarati & hausa & hindi &  igbo & indonesian\\
\bottomrule

\end{tabular}
}
\end{center}
\caption{ XL-Sum$^{ar,en,ja,zh}$ test  set results  for all languages. We evaluate using the best performing model on the four source languages on the validation set.
}
\label{tab:xlsum-arenjazh-test-results}
\end{table*}

\begin{table*}[]
\begin{center}
\def\arraystretch{0.87}
\resizebox{0.99\textwidth}{!}{%
\addtolength{\tabcolsep}{-0.5pt}
\begin{tabular}{llrrrrrrr}
\toprule
       & lang &    amharic & arabic & azerbaijani & bengali & burmese & chinese\_simplified & chinese\_traditional \\
        &  &    Rg$_1$ / Rg$_2$ / Rg$_L$ &   Rg$_1$ / Rg$_2$ / Rg$_L$  &   Rg$_1$ / Rg$_2$ / Rg$_L$  &   Rg$_1$ / Rg$_2$ / Rg$_L$  &   Rg$_1$ / Rg$_2$ / Rg$_L$  &   Rg$_1$ / Rg$_2$ / Rg$_L$  &   Rg$_1$ / Rg$_2$ / Rg$_L$  \\
\midrule
\multirow{2}{*}{small} & \model & 12.3 / 1.7 / 11.7 & 	16.9 / 3.3 / 15.4 & 	13.7 / 3.6 / 11.8 & 	8.5 / 2.1 / 7.7 & 	17.6 / 3.7 / 15.8 & 	4.0 / 1.2 / 3.7 & 	5.3 / 2.0 / 4.9  \\
 & \shared & 13.6 / 1.2 / 12.9 & 	15.2 / 0.6 / 14.9 & 	13.9 / 1.7 / 12.7 & 	9.8 / 0.4 / 9.6 & 	18.3 / 0.6 / 18.0 & 	5.5 / 1.2 / 5.3 & 	5.9 / 1.7 / 5.7 \\
\midrule
\midrule
\multirow{2}{*}{base} & \model &  16.6 / 5.2 / 14.4 & 	19.3 / 4.5 / 17.1 & 	17.5 / 5.2 / 14.6 & 	15.1 / 5.2 / 12.7 & 	22.0 / 4.9 / 19.6 & 	9.2 / 2.8 / 8.3 & 	8.3 / 2.6 / 7.5  \\
 & \shared &   13.7 / 1.7 / 13.1 & 	15.8 / 1.3 / 15.1 & 	15.3 / 2.5 / 13.5 & 	10.5 / 0.8 / 10.0 & 	18.1 / 1.6 / 17.2 & 	6.7 / 1.8 / 6.4 & 	6.9 / 2.3 / 6.6  \\
\bottomrule

\\

\toprule
       & lang &     english & french & gujarati & hausa & hindi &  igbo & indonesian\\
\bottomrule

\end{tabular}
}
\end{center}
\caption{ XL-Sum$^{en}$ test  set results  for all languages. We evaluate using the best performing model on the English language of the validation set.
}
\label{tab:XLSUM_en_test_results}
\end{table*}

\begin{table*}[]
\begin{center}
\def\arraystretch{0.87}
\resizebox{0.99\textwidth}{!}{%
\addtolength{\tabcolsep}{-0.5pt}
\begin{tabular}{cllrrrrrrr}

\toprule
  &     &  &    amharic & arabic & azerbaijani & bengali & burmese & chinese\_simplified & chinese\_traditional \\
   &     &  &    Rg$_1$ / Rg$_2$ / Rg$_L$ &   Rg$_1$ / Rg$_2$ / Rg$_L$  &   Rg$_1$ / Rg$_2$ / Rg$_L$  &   Rg$_1$ / Rg$_2$ / Rg$_L$  &   Rg$_1$ / Rg$_2$ / Rg$_L$  &   Rg$_1$ / Rg$_2$ / Rg$_L$  &   Rg$_1$ / Rg$_2$ / Rg$_L$  \\
\midrule
\multirow{2}{*}{\rotatebox[origin=c]{90}{Small}}&\multirow{1}{*}{\model} & s7 &  12.5 / 1.8 / 11.9 & 	17.2 / 3.3 / 15.7 & 	14.0 / 4.0 / 12.1 & 	8.7 / 2.2 / 7.9 & 	17.8 / 3.6 / 16.2 & 	3.9 / 1.2 / 3.6 & 	5.4 / 2.0 / 4.9  \\
\cline{2-10}
&\multirow{1}{*}{\shared} & s1 &  13.6 / 1.2 / 13.0 & 	15.3 / 0.6 / 15.0 & 	14.0 / 1.9 / 12.7 & 	9.5 / 0.3 / 9.4 & 	18.1 / 0.5 / 17.8 & 	5.6 / 1.2 / 5.4 & 	5.8 / 1.6 / 5.6 \\  
\midrule
\multirow{4}{*}{\rotatebox[origin=c]{90}{Base}}&\multirow{2}{*}{\model} & s1 & 	12.6 / 0.2 / 12.4 & 	15.0 / 0.3 / 14.7 & 	13.9 / 1.4 / 12.5 & 	9.5 / 0.2 / 9.3 & 	17.8 / 0.2 / 17.6 & 	5.3 / 0.5 / 5.1 & 	5.0 / 0.5 / 4.9    \\ 
& & s7 &         16.3 / 5.0 / 14.2 & 	19.8 / 4.7 / 17.4 & 	17.7 / 5.4 / 14.7 & 	15.0 / 5.0 / 12.5 & 	22.0 / 4.8 / 19.5 & 	9.1 / 2.7 / 8.1 & 	8.1 / 2.5 / 7.3   \\
\cline{2-10}
&\multirow{2}{*}{\shared} & s1 & 	13.8 / 1.6 / 13.0 & 	16.0 / 1.3 / 15.3 & 	15.4 / 2.9 / 13.6 & 	9.9 / 0.6 / 9.5 & 	17.8 / 1.7 / 16.9 & 	6.6 / 1.8 / 6.3 & 	6.8 / 2.2 / 6.5   \\ 
& & s7 &         13.7 / 1.4 / 13.1 & 	16.7 / 1.3 / 16.0 & 	16.7 / 3.6 / 14.7 & 	9.9 / 0.5 / 9.6 & 	18.2 / 1.4 / 17.4 & 	6.4 / 1.6 / 6.2 & 	6.0 / 1.6 / 5.8     \\
\bottomrule

\end{tabular}
}
\end{center}
\caption{ XL-Sum$^{en}$ validation  results for each of the languages. We report results of different combinations of freezing the model. 
}
\label{tab:XLSUM_en_dev_results}
\end{table*}

\begin{table*}[]
\begin{center}
\def\arraystretch{0.87}
\resizebox{0.99\textwidth}{!}{%
\addtolength{\tabcolsep}{-0.5pt}
\begin{tabular}{llrrrrrrr}
\toprule
       &  &    amharic & arabic & azerbaijani & bengali & burmese & chinese\_simplified & chinese\_traditional \\
        &  &    Rg$_1$ / Rg$_2$ / Rg$_L$ &   Rg$_1$ / Rg$_2$ / Rg$_L$  &   Rg$_1$ / Rg$_2$ / Rg$_L$  &   Rg$_1$ / Rg$_2$ / Rg$_L$  &   Rg$_1$ / Rg$_2$ / Rg$_L$  &   Rg$_1$ / Rg$_2$ / Rg$_L$  &   Rg$_1$ / Rg$_2$ / Rg$_L$  \\
\midrule
\multirow{2}{*}{small} & \model &  \textbf{37.5} / \textbf{19.2} / 27.9 & 	36.6 / 16.6 / 28.5 & 	\textbf{28.3} / \textbf{13.4} / \textbf{23.5} & 	\textbf{35.4} / \textbf{19.5} / \textbf{28.0} & 	43.3 / 23.1 / \textbf{34.1} & 	36.7 / 17.9 / 31.7 & 	37.9 / 19.6 / 32.1 \\

 & \shared &  37.2 / \textbf{19.2} / \textbf{28.1} & 	\textbf{36.9} / \textbf{16.7} / \textbf{28.9} & 	27.8 / 13.2 / 23.1 & 	34.5 / 18.8 / 27.5 & 	\textbf{43.4} / \textbf{23.4} / 34.0 & 	\textbf{37.1} / \textbf{18.1} / \textbf{32.1} & 	\textbf{38.1} / \textbf{19.8} / \textbf{32.3} \\
\midrule
\midrule
\multirow{2}{*}{base} & \model &  \textbf{42.9} / \textbf{25.1} / \textbf{33.0} & 	\textbf{41.9} / \textbf{21.5} / \textbf{33.0} & 	\textbf{33.6} / \textbf{18.3} / \textbf{28.4} & 	\textbf{40.7} / \textbf{24.7} / \textbf{32.4} & 	\textbf{50.0} / \textbf{29.9} / \textbf{39.6} & 	\textbf{42.2} / \textbf{22.3} / \textbf{36.6} & 	\textbf{43.0} / \textbf{24.0} / \textbf{36.7} \\

 & \shared &  40.1 / 22.0 / 30.0 & 	39.7 / 19.4 / 30.9 & 	30.3 / 15.3 / 25.2 & 	37.9 / 21.7 / 29.5 & 	45.7 / 25.3 / 35.6 & 	39.8 / 20.4 / 34.3 & 	40.8 / 22.3 / 34.7  \\
\bottomrule

\\

\toprule
       &  &    english & french & gujarati & hausa & hindi &  igbo & indonesian  \\
\bottomrule

\end{tabular}
}
\end{center}
\caption{ Test set results for XL-Sum in the multisource setup. We evaluate using the model which performed best on all the languages in the validation set.
}
\label{tab:XLSUM_test_results}
\end{table*}

\begin{table*}[]
\begin{center}
\def\arraystretch{0.87}
\resizebox{0.99\textwidth}{!}{%
\addtolength{\tabcolsep}{-0.5pt}
\begin{tabular}{cllrrrrrrr}
\toprule
    &   & lang &    amharic & arabic & azerbaijani & bengali & burmese & chinese\_simplified & chinese\_traditional \\
  &      &  &    Rg$_1$ / Rg$_2$ / Rg$_L$ &   Rg$_1$ / Rg$_2$ / Rg$_L$  &   Rg$_1$ / Rg$_2$ / Rg$_L$  &   Rg$_1$ / Rg$_2$ / Rg$_L$  &   Rg$_1$ / Rg$_2$ / Rg$_L$  &   Rg$_1$ / Rg$_2$ / Rg$_L$  &   Rg$_1$ / Rg$_2$ / Rg$_L$  \\
  
  \midrule
\multirow{2}{*}{\rotatebox[origin=c]{90}{Small}}&\multirow{1}{*}{\model} & s7 & 	29.1 / 13.0 / 22.2 & 	37.7 / 17.3 / 29.6 & 	19.0 / 7.2 / 15.9 & 	27.6 / 12.8 / 22.3 & 	38.3 / 17.7 / 29.7 & 	38.3 / 19.1 / 33.2 & 	39.8 / 21.4 / 34.0   \\ 
 \cline{2-10}
&\multirow{1}{*}{\shared} & s1 & 13.0 / 3.1 / 12.0 & 	37.6 / 17.3 / 29.4 & 	13.4 / 2.6 / 12.1 & 	7.4 / 1.1 / 7.2 & 	11.9 / 1.9 / 11.4 & 	38.5 / 19.5 / 33.4 & 	39.9 / 21.6 / 34.1 \\ 
\midrule
  
\midrule
\multirow{4}{*}{\rotatebox[origin=c]{90}{Base}}&\multirow{2}{*}{\model} & s1 & 	9.4 / 0.3 / 9.3 & 	41.2 / 20.9 / 31.9 & 	12.3 / 1.8 / 11.2 & 	8.3 / 0.4 / 8.1 & 	15.1 / 0.3 / 14.9 & 	64.1 / 51.2 / 60.8 & 	62.9 / 49.8 / 58.8  \\ 
& & s7 &         34.0 / 16.6 / 25.3 & 	42.4 / 22.3 / 33.5 & 	26.3 / 11.0 / 21.1 & 	34.5 / 18.0 / 26.7 & 	40.1 / 19.0 / 30.9 & 	48.4 / 29.6 / 43.2 & 	49.4 / 31.3 / 43.6  \\ 
 \cline{2-10}
&\multirow{2}{*}{\shared} & s1 & 	14.9 / 4.2 / 13.6 & 	40.4 / 20.1 / 31.5 & 	15.9 / 4.4 / 14.1 & 	11.4 / 3.1 / 10.6 & 	16.5 / 4.1 / 15.1 & 	44.3 / 26.2 / 39.3 & 	45.4 / 27.8 / 39.7  \\ 
& & s7 &         10.3 / 3.4 / 9.4 & 	40.5 / 20.1 / 31.8 & 	12.3 / 4.2 / 10.9 & 	6.3 / 2.0 / 5.9 & 	22.3 / 6.3 / 19.9 & 	41.5 / 21.9 / 36.0 & 	42.7 / 24.2 / 36.8  \\
\bottomrule

\end{tabular}
}
\end{center}
\caption{ XL-Sum$^{ar,en,ja,zh}$ validation set results for all languages. We report results for the different freezing configurations.
}
\label{tab:XLSUM_highres_dev_results}
\end{table*}
 
\begin{table*}[]
\begin{center}
\tiny
\def\arraystretch{0.82}
\resizebox{0.82\textwidth}{!}{%
\addtolength{\tabcolsep}{-0.2pt}
\begin{tabular}{cllcccccccccccc}
\toprule
& & & af\_ZA  &  am\_ET  &  ar\_SA  &  az\_AZ  &  bn\_BD  &  cy\_GB  &  da\_DK  &  de\_DE  &  el\_GR  &  en\_US  \\
\midrule
\multirow{10}{*}{\rotatebox[origin=c]{90}{Small}}&\multirow{5}{*}{\rotatebox[origin=c]{90}{\model}} 
& s1 & 27.4   &  5.3   &  14.0   &  12.9   &  9.7   &  8.8   &  31.1   &  30.8   &  21.1   &  69.4   \\
& & s6 & 35.6   &  6.2   &  19.2   &  16.9   &  11.4   &  7.8   &  45.4   &  41.4   &  28.6   &  72.8   \\
& & s7 & 39.5   &  5.6   &  20.9   &  18.3   &  11.9   &  10.0   &  46.6   &  42.6   &  31.4   &  73.0   \\
& & s10 & 38.6   &  6.9   &  19.9   &  19.0   &  12.4   &  9.2   &  47.0   &  41.8   &  30.2   &  72.2   \\
& & s14 & 38.6   &  6.6   &  20.1   &  18.4   &  13.6   &  10.3   &  46.7   &  43.5   &  30.1   &  73.2   \\
\cline{2-15}
\vspace{-0.7em}\\
&\multirow{5}{*}{\rotatebox[origin=c]{90}{\shared}} 
& s1 & 25.6   &  5.5   &  16.6   &  12.8   &  9.8   &  5.6   &  35.1   &  31.0   &  22.8   &  71.0   \\
\bottomrule
\end{tabular}
}
\end{center}
\caption{MASSIVE Exact Match (EM) dev accuracies of all models and settings for all languages.}
\label{tab:MASSIVE_DEV_RESULTS}
\end{table*}

\begin{table*}[]
\begin{center}
\def\arraystretch{0.87}
\resizebox{0.99\textwidth}{!}{%
\addtolength{\tabcolsep}{-0.5pt}

\begin{tabular}{cllrrrrrrr}
\toprule
  &     & lang &    english & french & gujarati & hausa & hindi &  igbo & indonesian  \\
  &      &  &    Rg$_1$ / Rg$_2$ / Rg$_L$ &   Rg$_1$ / Rg$_2$ / Rg$_L$  &   Rg$_1$ / Rg$_2$ / Rg$_L$  &   Rg$_1$ / Rg$_2$ / Rg$_L$  &   Rg$_1$ / Rg$_2$ / Rg$_L$  &   Rg$_1$ / Rg$_2$ / Rg$_L$  &   Rg$_1$ / Rg$_2$ / Rg$_L$  \\
\midrule
\multirow{10}{*}{\rotatebox[origin=c]{90}{Small}}&\multirow{5}{*}{\rotatebox[origin=c]{90}{\model}} & s1 & 	38.5 / 30.0 / 33.5 & 	38.1 / 29.3 / 31.5 & 	33.5 / 27.0 / 31.2 & 	41.9 / 32.6 / 36.0 & 	39.0 / 31.2 / 35.7 & 	43.0 / 31.8 / 34.5 & 	38.0 / 30.0 / 33.8 \\ 
& & s6 & 	36.1 / 28.4 / 32.7 & 	38.0 / 29.4 / 32.4 & 	32.0 / 26.0 / 31.9 & 	39.4 / 30.9 / 36.1 & 	36.8 / 29.5 / 35.3 & 	38.2 / 28.8 / 34.6 & 	35.6 / 28.1 / 33.2 \\ 
 && s7 & 	37.3 / 29.1 / 33.5 & 	37.9 / 29.1 / 33.0 & 	32.8 / 26.5 / 31.8 & 	40.4 / 31.4 / 37.1 & 	38.1 / 30.4 / 35.8 & 	41.0 / 30.8 / 35.0 & 	36.5 / 28.9 / 34.1 \\ 
 && s10 & 	36.4 / 28.5 / 32.4 & 	37.9 / 29.5 / 32.0 & 	32.5 / 26.5 / 31.2 & 	39.4 / 30.7 / 35.6 & 	36.9 / 29.6 / 34.9 & 	38.5 / 29.3 / 33.8 & 	35.9 / 28.3 / 32.9 \\ 
 & &s14 & 	36.7 / 28.7 / 33.3 & 	38.3 / 29.5 / 33.1 & 	32.8 / 26.7 / 31.5 & 	40.4 / 31.4 / 36.6 & 	37.8 / 30.2 / 35.7 & 	41.3 / 31.2 / 34.9 & 	36.3 / 28.6 / 34.1 \\ 
							\cline{2-10}
&\multirow{5}{*}{\rotatebox[origin=c]{90}{\shared}}  & s1 & 	38.3 / 29.8 / 31.8 & 	38.3 / 29.6 / 30.9 & 	33.0 / 26.5 / 29.1 & 	41.3 / 32.0 / 34.2 & 	38.6 / 30.8 / 33.7 & 	42.8 / 31.6 / 31.9 & 	37.7 / 30.0 / 32.1 \\
\bottomrule
\end{tabular}
}
\end{center}
\caption{ XL-Sum results . Different configurations of freezing.
}
\label{tab:XLSUM_dev_results_english_etc}
\end{table*}

\begin{table*}[]
\begin{center}
\def\arraystretch{0.87}
\resizebox{0.99\textwidth}{!}{%
\addtolength{\tabcolsep}{-0.5pt}

\begin{tabular}{cllrrrrrrr}
\toprule
      
    &   & lang &   japanese & korean & kyrgyz & marathi & nepali & pashto & persian  \\
     &   &  &    Rg$_1$ / Rg$_2$ / Rg$_L$ &   Rg$_1$ / Rg$_2$ / Rg$_L$  &   Rg$_1$ / Rg$_2$ / Rg$_L$  &   Rg$_1$ / Rg$_2$ / Rg$_L$  &   Rg$_1$ / Rg$_2$ / Rg$_L$  &   Rg$_1$ / Rg$_2$ / Rg$_L$  &   Rg$_1$ / Rg$_2$ / Rg$_L$  \\
        \midrule
      \multirow{10}{*}{\rotatebox[origin=c]{90}{Small}}&  \multirow{5}{*}{\rotatebox[origin=c]{90}{\model}} & s1 & 	42.2 / 34.3 / 39.5 & 	38.6 / 33.2 / 36.7 & 	26.7 / 21.1 / 23.8 & 	32.6 / 26.5 / 30.2 & 	39.8 / 32.6 / 36.5 & 	44.4 / 34.1 / 38.6 & 	43.1 / 33.8 / 37.3 \\ 
 && s6 & 	39.1 / 32.5 / 39.9 & 	37.9 / 32.3 / 38.8 & 	25.6 / 21.2 / 25.3 & 	30.6 / 25.4 / 29.8 & 	36.8 / 30.1 / 36.9 & 	42.7 / 33.1 / 38.4 & 	41.4 / 32.6 / 36.8 \\ 
 && s7 & 	41.5 / 34.3 / 41.1 & 	39.3 / 34.0 / 39.2 & 	26.4 / 21.6 / 25.4 & 	31.9 / 26.3 / 30.6 & 	38.3 / 31.3 / 36.7 & 	43.7 / 33.9 / 38.9 & 	42.0 / 32.7 / 37.6 \\ 
 & &s10 & 	39.8 / 32.9 / 39.5 & 	38.3 / 32.9 / 38.7 & 	26.3 / 21.8 / 25.2 & 	30.9 / 25.4 / 29.7 & 	37.3 / 30.4 / 36.3 & 	42.5 / 33.1 / 38.0 & 	41.6 / 32.7 / 36.6 \\ 
 & &s14 & 	41.4 / 34.0 / 41.1 & 	39.4 / 34.0 / 39.2 & 	26.1 / 21.7 / 25.4 & 	32.4 / 26.6 / 30.5 & 	38.3 / 31.1 / 37.3 & 	43.4 / 33.6 / 38.5 & 	41.8 / 32.7 / 37.4 \\ 
							\cline{2-10}
&\multirow{5}{*}{\rotatebox[origin=c]{90}{\shared}} & s1 & 	42.9 / 34.8 / 37.9 & 	38.6 / 32.9 / 35.1 & 	26.4 / 21.4 / 22.1 & 	32.9 / 26.9 / 28.7 & 	39.0 / 31.6 / 34.5 & 	44.2 / 34.0 / 36.2 & 	42.8 / 33.5 / 35.8 \\
\bottomrule
\end{tabular}
}
\end{center}
\caption{ Validation set results for XL-Sum in the multisource setup for languages English, French, Gujarati, Hausa, Hindi, Igbo, and Indonesian. We report results for the different freezing configurations.
}
\label{tab:XLSUM_dev_results_japanese_etc}
\end{table*}

\begin{table*}[]
\begin{center}
\def\arraystretch{0.87}
\resizebox{0.99\textwidth}{!}{%
\addtolength{\tabcolsep}{-0.5pt}
\begin{tabular}{cllrrrrrrr}
\toprule
   &    & lang &  portuguese & punjabi & russian & scottish\_gaelic & serbian\_cyrillic & serbian\_latin & sinhala  \\
    &    &  &    Rg$_1$ / Rg$_2$ / Rg$_L$ &   Rg$_1$ / Rg$_2$ / Rg$_L$  &   Rg$_1$ / Rg$_2$ / Rg$_L$  &   Rg$_1$ / Rg$_2$ / Rg$_L$  &   Rg$_1$ / Rg$_2$ / Rg$_L$  &   Rg$_1$ / Rg$_2$ / Rg$_L$  &   Rg$_1$ / Rg$_2$ / Rg$_L$  \\
        \midrule
    \multirow{10}{*}{\rotatebox[origin=c]{90}{Small}}&    \multirow{5}{*}{\rotatebox[origin=c]{90}{\model}} & s1 & 	40.6 / 30.4 / 33.5 & 	39.7 / 29.0 / 33.3 & 	33.1 / 25.5 / 28.8 & 	37.1 / 28.1 / 30.9 & 	30.5 / 23.1 / 25.6 & 	30.4 / 23.0 / 26.1 & 	36.4 / 29.2 / 31.9 \\ 
 && s6 & 	38.6 / 29.0 / 32.9 & 	37.7 / 28.2 / 33.7 & 	31.0 / 23.9 / 27.9 & 	34.8 / 27.4 / 32.2 & 	29.0 / 22.5 / 27.3 & 	23.6 / 19.1 / 26.4 & 	35.2 / 28.3 / 34.7 \\ 
 && s7 & 	39.5 / 29.3 / 33.7 & 	38.8 / 28.8 / 33.9 & 	31.7 / 24.4 / 28.8 & 	36.3 / 28.2 / 32.1 & 	29.4 / 22.7 / 27.9 & 	26.1 / 20.5 / 27.0 & 	35.5 / 28.7 / 35.0 \\ 
 && s10 & 	38.8 / 29.0 / 32.7 & 	37.7 / 28.3 / 33.2 & 	31.1 / 23.9 / 27.0 & 	35.3 / 27.9 / 32.4 & 	29.2 / 22.5 / 26.7 & 	23.5 / 18.8 / 25.5 & 	35.5 / 28.4 / 34.2 \\ 
 && s14 & 	39.4 / 29.3 / 33.5 & 	38.3 / 28.3 / 33.5 & 	31.0 / 23.9 / 28.7 & 	35.3 / 27.6 / 32.1 & 	29.2 / 22.6 / 27.9 & 	24.4 / 19.3 / 26.9 & 	35.0 / 28.1 / 34.3 \\ 
							\cline{2-10}
&\multirow{5}{*}{\rotatebox[origin=c]{90}{\shared}} & s1 & 	40.4 / 30.2 / 32.1 & 	39.7 / 29.1 / 31.3 & 	32.2 / 24.8 / 26.9 & 	36.9 / 27.9 / 28.3 & 	30.4 / 22.9 / 24.4 & 	30.2 / 23.0 / 24.6 & 	35.8 / 27.9 / 30.4 \\
\bottomrule
\end{tabular}
}
\end{center}
\caption{ Validation set results for XL-Sum in the multisource setup for languages Portuguese, Punjabi, Russian, Scottish Gaelic, Serbian, and Sinhala. We report results for the different freezing configurations.
}
\label{tab:XLSUM_dev_results_portugues_etc}
\end{table*}

\begin{table*}[]
\begin{center}
\def\arraystretch{0.87}
\resizebox{0.99\textwidth}{!}{%
\addtolength{\tabcolsep}{-0.5pt}
\begin{tabular}{cllrrrrrrr}
\toprule
   &    & lang &   somali & spanish & swahili & tamil & telugu &  thai & turkish \\
    &    &  &    Rg$_1$ / Rg$_2$ / Rg$_L$ &   Rg$_1$ / Rg$_2$ / Rg$_L$  &   Rg$_1$ / Rg$_2$ / Rg$_L$  &   Rg$_1$ / Rg$_2$ / Rg$_L$  &   Rg$_1$ / Rg$_2$ / Rg$_L$  &   Rg$_1$ / Rg$_2$ / Rg$_L$  &   Rg$_1$ / Rg$_2$ / Rg$_L$  \\
\midrule
\multirow{10}{*}{\rotatebox[origin=c]{90}{Small}}&\multirow{5}{*}{\rotatebox[origin=c]{90}{\model}} & s1 & 	38.4 / 28.3 / 30.6 & 	31.3 / 23.5 / 27.3 & 	38.9 / 30.0 / 33.3 & 	31.8 / 26.6 / 31.2 & 	29.0 / 23.6 / 27.5 & 	31.5 / 24.8 / 28.2 & 	33.2 / 27.8 / 32.0 \\ 
& & s6 & 	37.0 / 27.4 / 30.8 & 	32.2 / 24.2 / 27.1 & 	37.7 / 28.8 / 33.7 & 	28.9 / 24.2 / 31.3 & 	26.8 / 22.1 / 27.9 & 	31.0 / 24.8 / 29.3 & 	30.3 / 25.3 / 31.5 \\ 
 && s7 & 	37.5 / 27.7 / 31.2 & 	30.8 / 23.1 / 27.6 & 	38.3 / 29.5 / 34.7 & 	30.8 / 26.0 / 32.2 & 	27.6 / 22.7 / 28.4 & 	31.8 / 25.3 / 29.8 & 	31.5 / 26.4 / 32.4 \\ 
 && s10 & 	36.8 / 27.2 / 30.2 & 	31.9 / 23.9 / 26.7 & 	37.9 / 28.9 / 33.6 & 	28.8 / 24.0 / 31.0 & 	26.9 / 22.2 / 27.3 & 	31.5 / 25.0 / 29.3 & 	30.5 / 25.6 / 31.0 \\ 
 && s14 & 	37.6 / 27.6 / 31.0 & 	31.1 / 23.3 / 27.5 & 	38.3 / 29.4 / 34.3 & 	30.4 / 25.6 / 32.1 & 	27.7 / 22.6 / 28.2 & 	31.8 / 25.1 / 29.6 & 	31.2 / 26.1 / 32.1 \\ 
							\cline{2-10}
&\multirow{5}{*}{\rotatebox[origin=c]{90}{\shared}} & s1 & 	38.0 / 28.0 / 29.0 & 	31.7 / 23.9 / 25.6 & 	38.5 / 29.7 / 31.7 & 	31.9 / 26.7 / 28.7 & 	28.5 / 23.2 / 25.3 & 	31.1 / 24.4 / 26.2 & 	32.9 / 27.5 / 30.2 \\
\bottomrule
\end{tabular}
}
\end{center}
\caption{Validation set results for XL-Sum in the multisource setup for languages Somali, Spanish, Swahili, Tamil, Telugu, Thai, and Turkish . We report results for the different freezing configurations.
}
\label{tab:XLSUM_dev_results_somali_etc}
\end{table*}

\begin{table*}[]
\begin{center}
\def\arraystretch{0.87}
\resizebox{0.99\textwidth}{!}{%
\addtolength{\tabcolsep}{-0.5pt}
\begin{tabular}{cllrrrrrr|r}
\toprule
    &   & lang &   ukrainian &  urdu & uzbek & vietnamese & welsh & yoruba & \textit{avg}\\
     &   &  &    Rg$_1$ / Rg$_2$ / Rg$_L$ &   Rg$_1$ / Rg$_2$ / Rg$_L$  &   Rg$_1$ / Rg$_2$ / Rg$_L$  &   Rg$_1$ / Rg$_2$ / Rg$_L$  &   Rg$_1$ / Rg$_2$ / Rg$_L$  &   Rg$_1$ / Rg$_2$ / Rg$_L$  &   Rg$_1$ / Rg$_2$ / Rg$_L$  \\
\midrule
\multirow{10}{*}{\rotatebox[origin=c]{90}{Small}}&\multirow{5}{*}{\rotatebox[origin=c]{90}{\model}} & s1 & 	32.2 / 25.3 / 29.2 & 	42.3 / 33.0 / 37.3 & 	28.4 / 23.5 / 26.6 & 	44.8 / 31.0 / 34.4 & 	41.1 / 31.0 / 34.0 & 	45.2 / 33.2 / 35.6  & \textbf{36.8 / 18.3 / 28.8}\\ 
 && s6 & 	30.1 / 23.6 / 28.6 & 	39.5 / 30.6 / 36.8 & 	26.8 / 22.1 / 27.1 & 	42.9 / 29.9 / 34.1 & 	37.4 / 28.3 / 33.5 & 	42.9 / 31.8 / 35.8  & 34.8 / 16.5 / 27.5\\ 
 && s7 & 	30.5 / 23.7 / 29.2 & 	41.1 / 32.0 / 37.4 & 	27.3 / 22.4 / 27.5 & 	43.8 / 30.3 / 34.8 & 	39.5 / 29.8 / 34.0 & 	44.0 / 32.6 / 36.6  & 35.8 / 17.4 / 28.2\\ 
 && s10 & 	30.1 / 23.7 / 28.1 & 	39.8 / 30.9 / 36.4 & 	26.9 / 22.2 / 26.9 & 	43.2 / 30.1 / 34.1 & 	37.7 / 28.4 / 33.5 & 	42.8 / 31.7 / 35.5  & 35.0 / 16.7 / 27.6\\ 
 && s14 & 	30.5 / 23.8 / 29.0 & 	40.5 / 31.5 / 37.3 & 	26.8 / 22.0 / 27.4 & 	43.7 / 30.3 / 34.4 & 	39.3 / 29.9 / 34.1 & 	43.5 / 32.3 / 36.2  & 35.6 / 17.2 / 28.0\\ 
						\cline{2-10}
&\multirow{5}{*}{\rotatebox[origin=c]{90}{\shared}} & s1 & 	31.7 / 25.0 / 27.3 & 	41.6 / 32.4 / 35.3 & 	28.2 / 23.1 / 24.4 & 	44.8 / 30.8 / 33.0 & 	40.4 / 30.4 / 31.8 & 	44.3 / 32.7 / 34.2  & 36.6 / 18.0 / 28.6\\
\bottomrule
\end{tabular}
}
\end{center}
\caption{ Validation set results for XL-Sum in the multisource setup for languages Ukrainian, Urdu, Uzbek, Vietnamese, Welsh, and Yoruba. We report results for the different freezing configurations.
}
\label{tab:XLSUM_dev_results_ukrainian_etc}
\end{table*}

\begin{table*}[]
\begin{center}
\def\arraystretch{0.87}
\resizebox{0.99\textwidth}{!}{%
\addtolength{\tabcolsep}{-0.5pt}
\begin{tabular}{ccl rrrrrrrrrrrrrrrrrrrrrr}
\toprule
 & & &   ar & bn & en & fi & id & ko & ru & sw & te & avg \\
 &  & cfg &    F1 \, /  EM &    F1 \, /  EM  &    F1 \, /  EM  &    F1 \, /  EM  &    F1 \, /  EM  &    F1 \, /  EM  &    F1 \, /  EM &    F1 \, /  EM  &    F1 \, /  EM & F1 / EM \\
 \midrule

  \multirow{10}{*}{\rotatebox[origin=c]{90}{Small}} & \multirow{5}{*}{\rotatebox[origin=c]{90}{\model}} & s1 & 51.6 / 34.9  & 30.4 / 16.8  & 62.7 / 51.4  & 46.9 / 32.1  & 49.9 / 33.6  & 26.6 / 19.6  & 46.3 / 28.6  &  34.4 / 23.8 & 31.5 / 22.1  & 42.3 / 29.2   \\
  & & s6 &  57.5 / 39.4  & 39.5 / 24.8  & 68.3 / 58.2  & 51.7 / 35.2  & 54.3 / 40.0  & 26.5 / 15.9  & 58.2 / 36.5  & 44.8 / 32.5  & 43.5 / 29.6  & 49.4 / 34.7  \\
& & s7 & 63.2 / 46.0  & 40.6 / 26.5  & 70.7 / 60.5  & 59.3 / 41.9  & 57.9 / 43.4  & 30.7 / 18.5  & 60.3 / 39.9  & 36.4 / 25.7  & 38.1 / 23.9  & 50.8 / 36.3  \\
& & s10 & 39.7 / 56.8 & 25.7 / 40.6 & 57.7 / 69.4 & 34.7 / 49.1 & 40.2 / 53.6 & 17.0 / 26.9 & 34.7 / 57.8 & 31.5 / 45.9 & 28.7 / 39.5 & 34.4 / 48.7  \\
& & s14 & 63.1 / 47.6   & 40.5 / 25.7  & 71.0 / 60.2  & 58.1 / 40.3  & 56.1 / 41.1  & 30.4 / 18.1  & 59.3 / 37.9  & 40.9 / 29.9  & 33.9 / 22.9  & 50.4 / 36.0  \\
 \cmidrule{2-13}
 
& \multirow{5}{*}{\rotatebox[origin=c]{90}{\shared}} & s1 &  45.6 / 30.2  & 25.5 / 14.2  & 61.7 / 50.0  &  45.1 / 29.9 & 49.7 / 33.8  & 27.1 / 20.3  & 50.8 / 35.0  & 33.1 / 21.8  & 18.5 / 13.6  & 39.7 / 27.6  \\
\bottomrule
\end{tabular}
}
\end{center}
\caption{ Results for the validation set of TyDiQA. We report results for the different configurations of freezing.
}
\label{tab:TYDIQA_dev_results}
\end{table*}

\end{document}